\documentclass[11pt]{article}

\usepackage{acl}
\usepackage{multirow}
\usepackage{microtype}

\usepackage{graphicx}

\usepackage{amsmath}
\usepackage{booktabs}
\usepackage{tabularx}
\usepackage{enumitem}
\usepackage{tcolorbox}

\usepackage[english,bidi=default]{babel}
\babelfont{rm}[
  Extension      = .otf,
  UprightFont    = *-regular,
  BoldFont       = *-bold,
  ItalicFont     = *-italic,
  BoldItalicFont = *-bolditalic
]{texgyretermes}
\babelprovide[import]{arabic}
\babelfont[*arabic]{rm}[
  Extension      = .ttf,
  UprightFont    = *-Regular,
  BoldFont       = *-Bold,
  ItalicFont     = *-Italic,
  BoldItalicFont = *-BoldItalic,
  Script         = Arabic
]{Amiri} 

\newcommand{\ar}[1]{\foreignlanguage{arabic}{#1}}

\newtcolorbox{promptbox}{
  colback=gray!5,
  colframe=gray!50,
  boxrule=0.5pt,
  left=6pt,
  right=6pt,
  top=4pt,
  bottom=4pt,
  fontupper=\small
}

\title{Understanding the Sociocultural Dimensions of Mental Health Discourse in Arabic-Language X Communities}

\newcommand*{\affaddr}[1]{#1} 
\newcommand*{\affmark}[1][*]{\textsuperscript{#1}}
\newcommand*{\email}[1]{\texttt{#1}}

\author{%
\bf{Amal Alqahtani \affmark[1], Rana Salama \affmark[2], and Mona Diab\affmark[3]}\\
\affaddr{\affmark[1]King Saud University, Riyadh, KSA}\\
\affaddr{\affmark[2]Faculty of Computers and Artificial Intelligence, Cairo University, Egypt}\\
\affaddr{\affmark[3]Carnegie Mellon University, Pittsburgh, Pennsylvania, USA}\\
\email{amaalqahtani@ksu.edu.sa, r.aref@fci-cu.edu.eg, mdiab@andrew.cmu.edu}\\}

\begin{document}
\maketitle

\begin{abstract}
Computational mental health research has predominantly centered on English-speaking populations, leaving Arabic-language discourse comparatively under-examined. We present an exploratory computational study of \textbf{8,147} tweets from \textbf{607} users classified by a GPT-4.1 personal-disclosure pipeline as likely lived-experience authors in three condition-specific Arabic-language X (formerly Twitter) Communities. We focus on discourse related to borderline personality disorder (BPD), bipolar disorder, and ADHD, and characterize community-associated linguistic patterns using a multi-domain cultural keyword framework. The results suggest that in this corpus, Bipolar tweets contain more religious and medical vocabulary, BPD tweets contain more relational, identity, and emotional-distress vocabulary, and ADHD tweets more often focus on practical symptoms and medication management. We treat these patterns as hypothesis-generating rather than confirmatory because the corpus is imbalanced across conditions, some subcorpora are temporally concentrated, and the keyword framework is an initial operationalization rather than a validated measurement instrument. The paper contributes a reusable LLM-assisted personal-disclosure pipeline and an exploratory cultural keyword framework for Arabic mental health discourse.
\end{abstract}

\section{Introduction}
In the Arab world, mental health discourse is strongly shaped by sociocultural norms. In particular, stigma associated with family honor and traditional religious interpretive frameworks continues to impede clinical help-seeking \citep{dardas2015stigma, zolezzi2018stigma}. Despite these barriers, condition-specific Arabic-language social media communities have emerged as important spaces for peer support and the exchange of lived experiences. Nevertheless, these communities remain largely underexplored in computational mental health research.
Existing computational mental health work has focused overwhelmingly on English, largely framing the problem as supervised classification of at-risk individuals \citep{coppersmith2014quantifying, de2013predicting}. Arabic NLP has been advanced by transformer-based models such as AraBERT \citep{antoun-etal-2020-arabert} and MARBERT \citep{abdul-mageed-etal-2021-arbert}, which establish strong baselines across Arabic NLP benchmarks, yet Arabic mental health NLP has paid comparatively less attention to culturally situated discourse characterization.

We adopt a \textit{characterization-oriented} approach grounded in Computational Social Science \citep{lazer2009computational}. By moving beyond the diagnostic paradigm, we prioritize a descriptive analysis of community-associated discourse, investigating how users articulate mental health experiences while avoiding clinical inferences about diagnosis or patient status.
We analyze \textbf{8,147} tweets from \textbf{607} users across three condition-specific X communities (BPD, Bipolar, ADHD) using a GPT-4.1 personal-disclosure pipeline validated against human annotators. 
\\Our main contributions include:
\begin{enumerate}[noitemsep, topsep=0pt]
\item \textbf{Dataset:} We introduce a multi-condition Arabic mental health corpus comprising 9,582 preprocessed posts, reduced to 8,147 posts after personal-disclosure filtering.\footnote{Available at: \url{https://github.com/amalqahtani/arabic-x-mental-health-discourse}.}
\item \textbf{Annotation Pipeline:} We develop a GPT-4.1-based tweet-level classification pipeline augmented with a reason-tag taxonomy and confidence scoring, and validate its outputs against a human-annotated gold standard.
\item \textbf{Discourse Analysis:} We conduct a comparative discourse analysis using circadian activity profiling, weighted log-odds, non-negative matrix factorization (NMF) topic modeling, and a six-domain cultural keyword framework.
\item \textbf{Empirical Findings:} We identify preliminary community-associated discourse patterns, including religious and medical vocabulary in Bipolar communities, identity and distress oriented language in BPD communities, and practical discussions of symptoms and medication management in ADHD communities. Given corpus imbalance and related methodological limitations, all analyses are interpreted as hypothesis-generating rather than confirmatory, and no between-community significance testing is performed.
\end{enumerate}

\section{Related Work}
\label{sec:related}
\vspace{-1mm}
\paragraph{Computational mental health on social media.} \citet{de2013predicting} showed that behavioral and linguistic signals in Twitter data predict depression onset. \citet{coppersmith2014quantifying} established a scalable self-reported diagnosis methodology and demonstrated condition-level linguistic differences across post-traumatic stress disorder (PTSD), depression, bipolar disorder, and social anxiety disorder (SAD). \citet{coppersmith2015adhd} extended this to ten conditions. More recently, \citet{yang-etal-2023-towards} explored LLM-generated explanations for mental health severity assessment, highlighting persistent challenges in grounding model outputs within clinically interpretable frameworks. Our work adapts the condition comparative paradigm to Arabic, using LLMs for personal disclosure annotation rather than classification, and reserving analysis for interpretable statistical methods. \vspace{-1mm} \paragraph{Arabic mental health, stigma, and cultural context.}
Mental health discourse in Arab societies is deeply shaped by social, religious, and cultural frameworks that influence how psychological distress is understood and discussed.
\citet{dardas2015stigma} discussed that mental illness stigma in Arab societies is closely intertwined with family honor norms and religious interpretive frameworks, while \citet{zolezzi2018stigma} confirmed these patterns in a systematic review spanning 33 studies. The theory of explanatory models \citep{kleinman1980patients}, which encompasses biomedical, spiritual, and relational interpretations of illness, provides the conceptual foundation for our analysis.
In Arab contexts, psychological distress may be interpreted through biomedical, religious, and supernatural frameworks, including explanations such as the evil eye (\textit{hasad}) or jinn possession (\textit{mass}), with religious or traditional healing sometimes considered alongside medical treatment \citep{eid2025somatic}. Emerging NLP evidence further reflects this explanatory pluralism. \citet{zaghouani-etal-2026-posts} found that religious and therapeutic vocabulary appear with comparable frequency in Arabic stress discourse, while \citet{ayash2025contextmentalqa} proposed that patient questions are frequently grounded in relational and faith-based reasoning that extends beyond conventional clinical taxonomies. Motivated by these findings, our work explicitly annotates sociocultural framing in Arabic mental health discourse, examining the co-occurrence of Social, Cultural, Religious, Medical, and Stigma dimensions across online mental health communities.
\vspace{-1mm}
\paragraph{LLM-Assisted Annotation for NLP.} Although traditional NLP classification pipelines rely on human-annotated ground truth, recent studies suggest that large language models (LLMs) can serve as effective annotators for complex and subjective tasks, with performance depending on the task domain, language, and prompting strategy \citep{gilardi2023chatgpt, ding-etal-2023-gpt}. In this work, we employ GPT-4.1 as the primary annotator for personal-disclosure identification (Section~\ref{sec:user-classification}) and assess reliability through human validation rather than assuming human-level equivalence. Collectively, these three bodies of work motivate our approach. We extend the condition-comparative framework of \citet{coppersmith2014quantifying} to Arabic, use LLMs for personal-disclosure annotation rather than clinical classification, operationalize Kleinman's \citeyear{kleinman1980patients} explanatory models computationally, and employ weighted log-odds, NMF topic modeling, and a cultural keyword framework to characterize sociocultural discourse.
\vspace{-2mm}
\section{Methodology}
\vspace{-2mm}
\subsection{Corpus Collection}
We collected Arabic-language posts from condition-specific X Communities on X (formerly Twitter) using publicly accessible platform data.

These X Communities are moderated spaces in which users join around shared interests and agree to community-specific participation rules prior to posting \footnote{\url{https://help.x.com/en/using-x/communities}}. Moderation practices vary across communities, ranging from professionally supervised spaces led by licensed psychologists to peer-supported groups. In Arabic-speaking contexts, where mental health conditions remain highly stigmatized \citep{dardas2015stigma, zolezzi2018stigma}, participation in condition-specific communities reflects meaningful engagement with mental health discourse. We therefore do not treat community membership as a diagnostic indicator; prior work has demonstrated that affiliation-based proxy signals perform poorly against clinical ground truth \citep{ernala2019methodological}. Instead, we use community structure as a pragmatic sampling frame for condition-relevant discourse \citep{abouwarda2024does}.
The resulting corpus consists of \textbf{10,091 tweets} collected from three condition-specific X Communities: BPD, Bipolar, and ADHD. Throughout this paper, these labels are capitalized when referring to the corresponding X Communities as data sources, whereas the associated clinical conditions (\textit{borderline personality disorder}, \textit{bipolar disorder}, and \textit{attention-deficit/hyperactivity disorder}) are written in lowercase.
Data were collected between March 31, 2022, and February 12, 2026. Each condition was represented by one Arabic-language X Community. Communities were selected based on four criteria: (1) an explicit focus on a specific mental health condition, (2) Arabic as the primary language of discourse, (3) active moderation, and (4) sustained member engagement. Additional community details are provided in (Appendix~\ref{appendix:communities}).
\vspace{-2mm}
\subsection{Pre-processing}
\begin{figure*}
    \centering
    \includegraphics[width=0.75\textwidth]{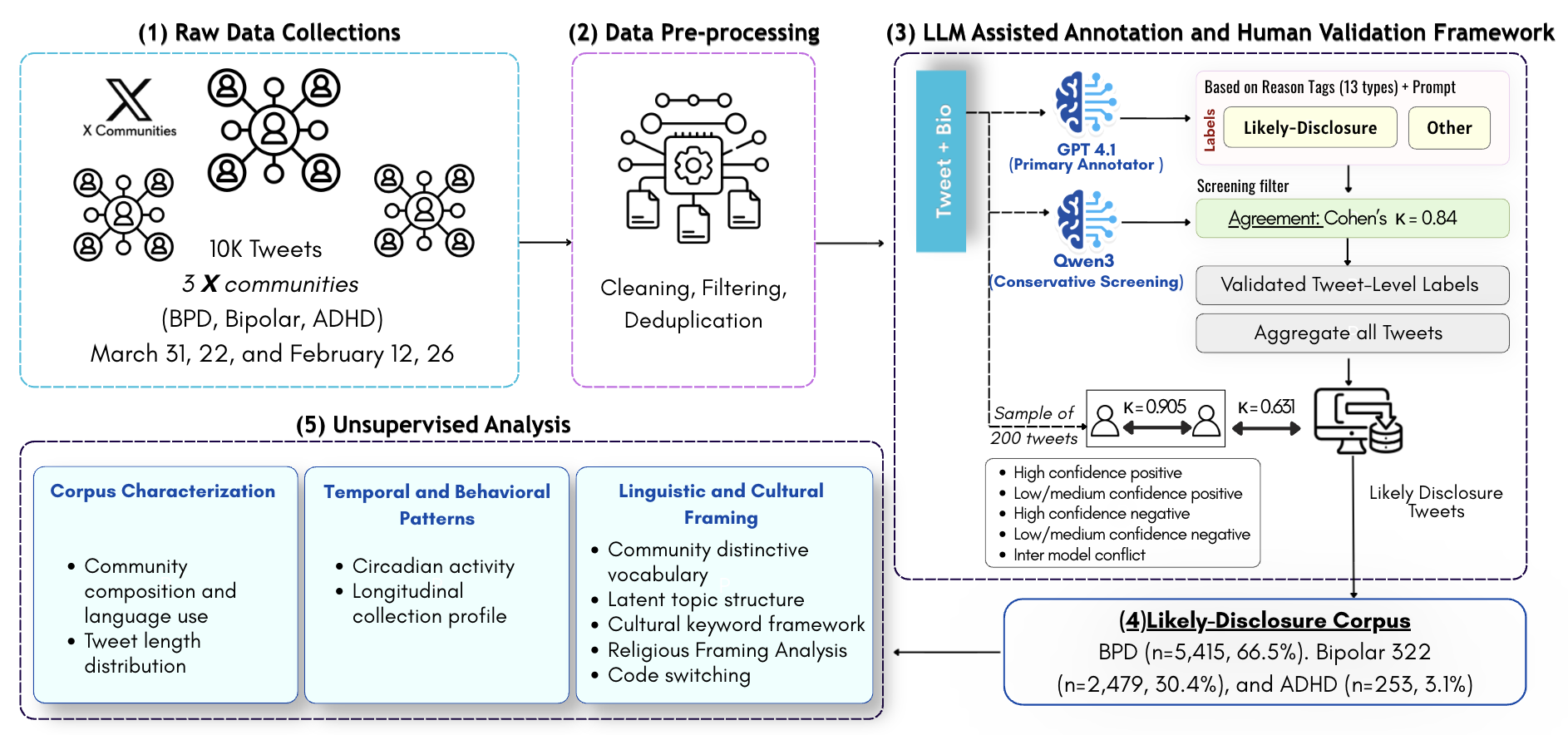}
    \caption{Overview of the computational pipeline. Exact collection period: March 31, 2022--February 12, 2026.}
    \label{fig:pipeline}
\end{figure*}

\vspace{-1mm}
From the initial set of 10,091 raw tweets, we removed URL-only posts, non-Arabic and non-English content, duplicate entries, and single-token tweets, resulting in a preprocessed corpus of \textbf{9,582} tweets produced by \textbf{1,286} unique users. The corpus is predominantly Arabic (98.4\%, $n$=9,428), with a small English component (1.2\%, $n$=116) and code-mixed undefined tweets (0.4\%, $n$=38). 
Figure~\ref{fig:pipeline} presents an overview of the complete data collection, annotation, and filtering pipeline.

\vspace{-2mm}
\subsection{LLM-Assisted Annotation and Human Validation Framework}
\label{sec:user-classification}
To identify users whose posts contain evidence of personal mental health disclosure, we applied a tweet-level personal-disclosure classification pipeline to all 9,582 preprocessed tweets from 1,286 unique users. Classification was performed using GPT-4.1\footnote{Model:\texttt{GPT-4.1-2025-04-14}} as the primary annotator, with Qwen3-235B-A22B \footnote {Model:\texttt{Qwen3-235B-A22B-Instruct-2507}} running the same prompt in parallel as a conservative screening model, both at temperature=0.0, max\_tokens=250. Each tweet was classified as either \textsc{Positive} (containing evidence of personal mental health disclosure) or \textsc{Negative} (containing no such evidence) with the user bio incorporated as supporting context. The prompt encodes explicit classification rules, a bio override rule, a conservative \textsc{Negative} default, and a 13 tag reason taxonomy (Appendix~\ref{appendix:prompt}). Disagreements between the two models flag ambiguous cases; inter-model agreement is reported in Appendix~\ref{appendix:validation}.
User-level aggregation was derived from tweet-level classifications using a deterministic priority-ordered aggregation procedure (Appendix~\ref{appendix:prompt-aggregation}). The resulting operational grouping was used exclusively for corpus filtering and downstream discourse analysis, and does not constitute a clinical or diagnostic categorization.
\\ \textbf{Likely personal-disclosure authors} are operationally defined as users for whom at least one tweet contains a personal-disclosure signal or whose bio includes self-identification language; this designation does not imply or establish a clinical diagnosis. \textbf{Other} users are those whose tweets contain no detected personal-disclosure signals and whose bios contain no self-identification indicators, including professionals, caregivers, and general community participants. Full aggregation rules are provided in Appendix~\ref{appendix:prompt-aggregation}. Pipeline reliability is assessed through a human validation study (Section~\ref{sec:human-validation} and Appendix~\ref{appendix:validation}). Under the GPT-4.1 primary annotation framework, \textbf{607} of 1,286 users (\textbf{47.2\%}) were labeled as likely personal-disclosure authors, while \textbf{679} users (\textbf{52.8\%}) showed no detectable personal-disclosure signals. Among the 607 users identified by GPT-4.1, \textbf{528} were also identified by Qwen3, whereas \textbf{79} represented GPT-positive/Qwen-negative disagreement cases. Users without detected personal-disclosure signals under GPT-4.1 were excluded from downstream analysis, resulting in a final dataset of \textbf{8,147} tweets from \textbf{607} users. In total, \textbf{1,435} tweets (\textbf{15.0\%}) were excluded together with their associated authors.
\vspace{-3mm}
\subsection{Human Validation}
\label{sec:human-validation}
\vspace{-2mm}
Two native Arabic-speaking annotators with prior experience in Arabic NLP independently annotated a stratified sample of 200 tweets using the annotation guidelines described in Appendix~\ref{appendix:validation}. At the tweet-level, GPT-4.1 labeled 47.4\% of tweets as positive and Qwen3 labeled 42.9\%, reaching consensus on 90.8\% ($\kappa = 0.84$; Appendix~\ref{appendix:validation}). As the results below show, human validation aligns more strongly with GPT-4.1, whose labels define the final corpus. \vspace{-2mm} \paragraph{Inter human agreement.} The two annotators agreed on 192 of 200 tweets (raw agreement 96.00\%), yielding $\kappa = 0.905$ which corresponds to almost perfect agreement  \citep{landis1977measurement}. All eight disagreements were directionally consistent, reflecting differences in thresholding for ambiguous positive cases rather than fundamentally conflicting interpretations. This high agreement ceiling suggests that the annotation task is well-defined and that the labeling guidelines are internally consistent. \vspace{-2mm} \paragraph{GPT-4.1 against human gold.}
Using the 192 mutually agreed tweets as the reference set, GPT-4.1 achieved $\kappa = 0.631$ (substantial agreement), with precision of 0.92, recall of 0.85, and $\mathrm{F}_1 = 0.88$ on the positive personal-disclosure class. This places GPT-4.1 approximately 0.27 $\kappa$ points below the inter-annotator agreement ceiling, supporting its use as the primary annotation model in this study. \vspace{-2mm} \paragraph{Qwen3-235B-A22B against human gold.} After excluding parse failures, Qwen3 achieved $\kappa = 0.329$ (fair agreement) against the human reference set, primarily due to lower recall (0.61) on the positive class. These results suggest that the high inter-model agreement between GPT-4.1 and Qwen3 ($\kappa = 0.84$) is largely driven by agreement on clear and predominantly negative cases rather than by near-human annotation reliability. Accordingly, Qwen3 was used as a conservative screening model whose disagreements with GPT-4.1 were treated as indicators of potentially ambiguous tweets, rather than as independent validation.
\vspace{-2mm}
\paragraph{Stratum-level results.}
GPT-4.1 achieved its highest agreement with human annotations on clear-label strata, with performance declining to 46\% agreement on low and medium confidence negative strata, where the conservative \textsc{Negative} default appears to suppress some genuine disclosures. Qwen3 agreed with human annotations on only 18\% of tweets within the inter-model conflict stratum, further indicating that such cases require human adjudication. Agreement also varied across communities, with the highest agreement observed for ADHD ($\kappa = 0.73$), followed by BPD ($\kappa = 0.66$), and the lowest for Bipolar ($\kappa \approx 0.49$). The lower agreement for Bipolar is consistent with the more indirect and metaphorical disclosure style observed in that community. Full per-stratum results and complete agreement tables are provided in Appendix~\ref{appendix:validation} (Tables~\ref{tab:iaa} and~\ref{tab:stratum_agreement}).
\vspace{-2mm}
\section{Exploratory Analyses of the Self-Disclosure-Filtered Corpus}
\label{sec:analyses}
\vspace{-2mm}
This section presents exploratory analyses conducted on the final corpus of \textbf{8,147} tweets produced by \textbf{607} users classified as likely personal-disclosure authors. The downstream analyses combine statistical and dictionary-based methods; however, topic interpretations and keyword-domain assignments involve researcher judgment and should therefore be regarded as exploratory analytical constructs rather than validated annotations. For clarity, we organize the analyses into three thematic groups: (i)~Corpus Characterization, (ii)~Temporal and Behavioral Patterns, and (iii)~Linguistic and Cultural Framing. These analyses include weighted log-odds distinctive vocabulary analysis, NMF topic modeling, cultural keyword profiling, religious framing analysis, and English code-switching analysis.
\vspace{-1mm}
\smallskip
\noindent\textbf{Note on ADHD subgroup size and statistical power.}
The ADHD subcorpus ($n$=\textbf{253}, \textbf{3.1}\% of the total corpus) is substantially smaller than the BPD ($n$=\textbf{5,415}) and Bipolar ($n$=\textbf{2,479}) subcorpora. As a result, ADHD-specific findings should be interpreted as preliminary and subject to greater uncertainty due to reduced statistical power. We retain ADHD in the main analysis for descriptive completeness, while treating ADHD-specific results as low-confidence.
\vspace{-3mm}
\subsection{Corpus Characterization}
\label{sec:corpus-char}
\begin{figure}[t]
  \centering
  \includegraphics[width=0.85\columnwidth]{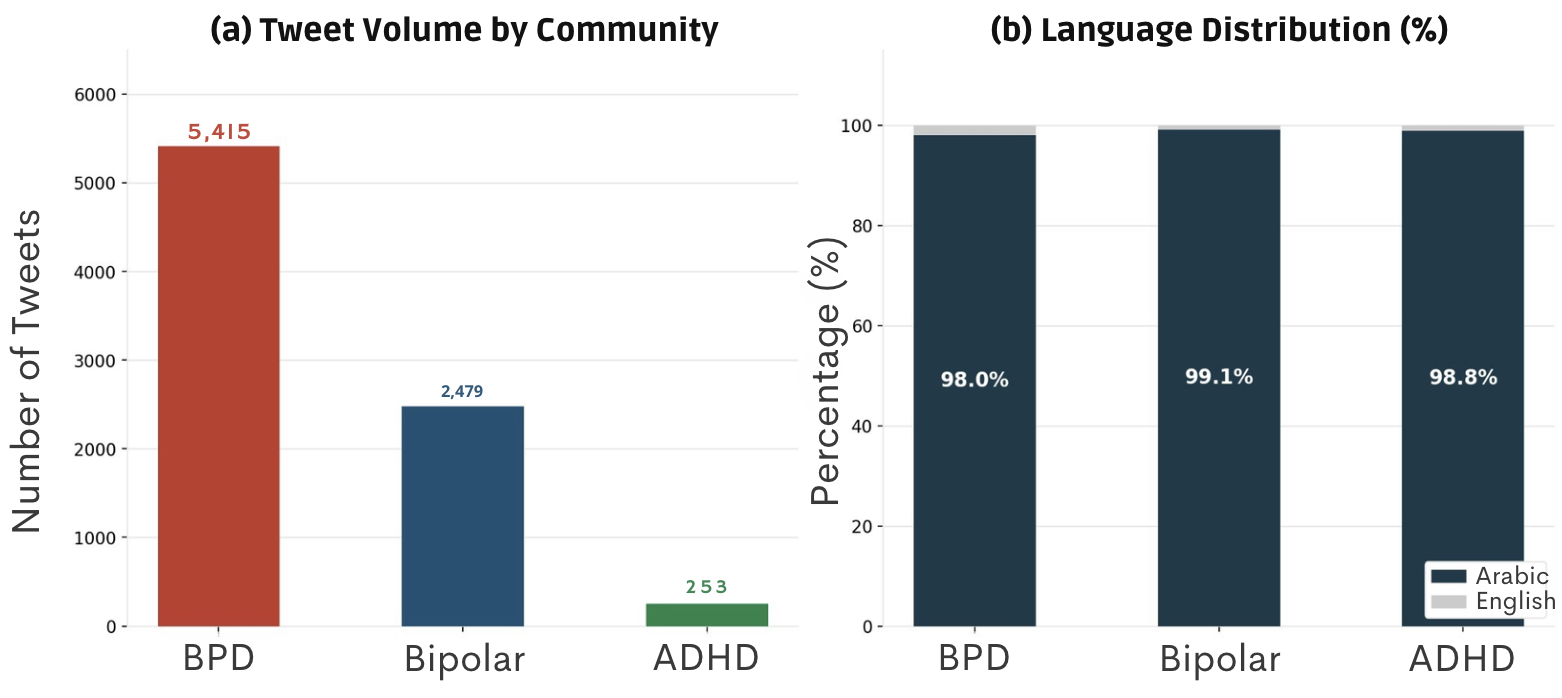}
  \caption{Tweet volume (left) and language distribution (right) per community. BPD dominates; all three communities are overwhelmingly Arabic-language.
  \vspace{-2mm}}
  \label{fig:corpus-overview}
\end{figure}

\vspace{-2mm}
\paragraph{Community composition and language use.}
Following the exclusion of users without detected personal-disclosure signals under the GPT-4.1 annotation framework, the filtered corpus comprises BPD ($n=\textbf{5,415}$; \textbf{66.5}\%), Bipolar ($n=\textbf{2,479}$; \textbf{30.4}\%), and ADHD ($n=\textbf{253}$; \textbf{3.1}\%). Discourse across all three communities is overwhelmingly Arabic-language (98.1--99.2\%; Figure~\ref{fig:corpus-overview}), distinguishing the corpus from predominantly English-centric social media mental health datasets and highlighting its value for under-resourced Arabic NLP research.

\begin{figure}[t]
  \centering
  \includegraphics[width=0.85\columnwidth]{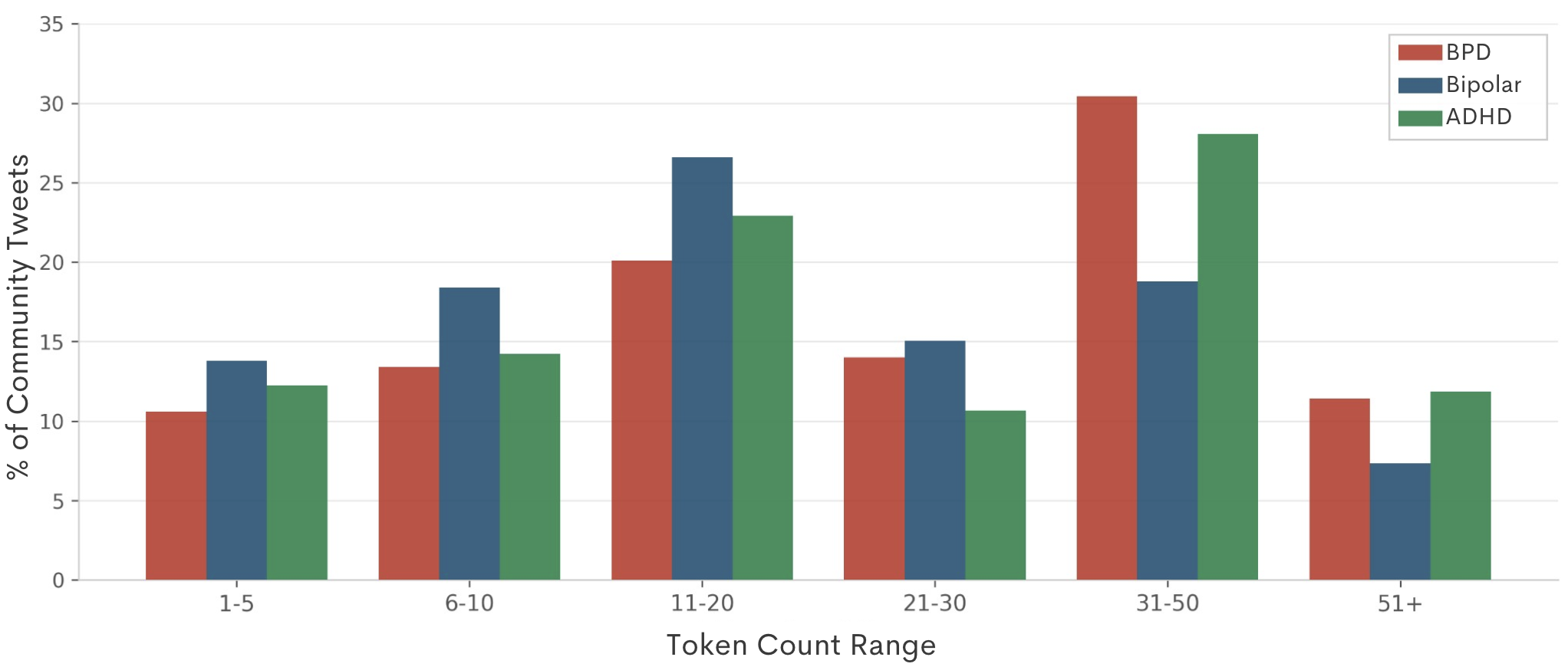}
  \caption{Normalized tweet length distributions. BPD and ADHD peak at 31--50 tokens; Bipolar at 11--20 tokens.
  \vspace{-3mm}
  }
  \label{fig:token-dist}
\end{figure}

\vspace{-2mm} \paragraph{Tweet length distribution.}
All three communities exhibit right-skewed token-length distributions (Figure~\ref{fig:token-dist}). Median tweet lengths are 25 tokens for BPD, 16 for Bipolar, and 21 for ADHD. BPD and ADHD discourse peaks within the 31 to 50 token range, whereas Bipolar discourse peaks earlier, within the 11 to 20 token range, consistent with shorter and more conversational interaction patterns. The pronounced long-tail distribution observed in BPD further suggests a higher prevalence of extended self-expression and help-seeking narratives.

\vspace{-3mm}
\subsection{Temporal and Behavioral Patterns}
\label{sec:temporal}
\begin{figure*}[t]
  \centering
  \includegraphics[width=0.70\textwidth]{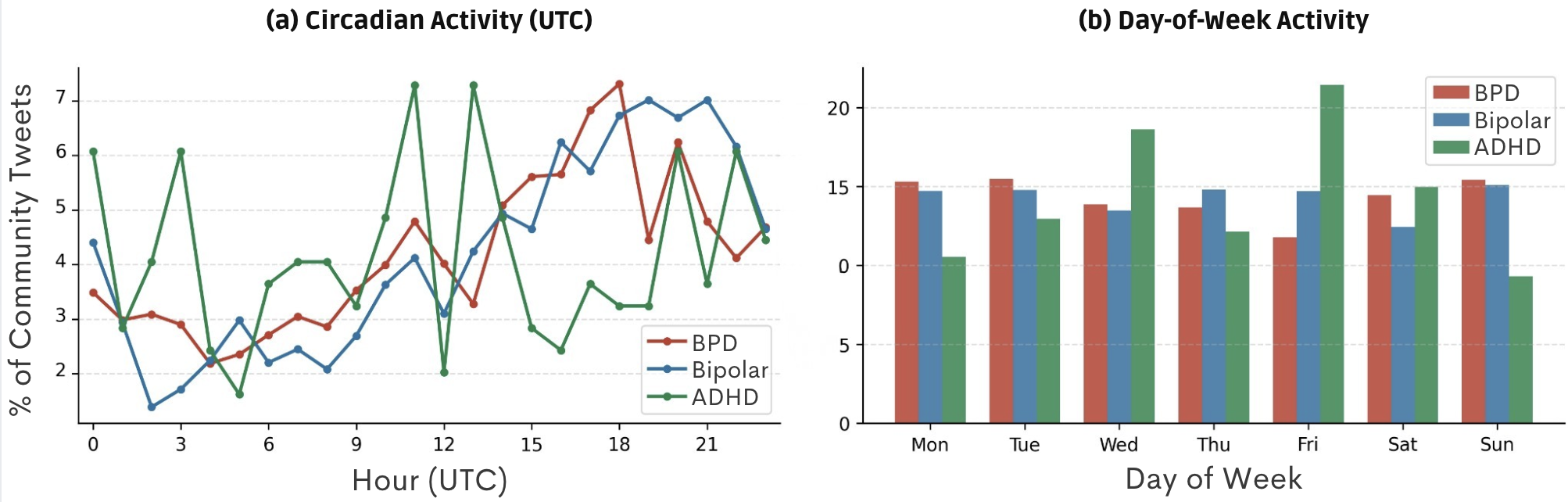}
  \caption{(a) Circadian (UTC) and (b) day-of-week tweet activity. BPD and Bipolar peak in the evening; ADHD at midday.
  \vspace{-2mm}
  }
  \label{fig:temporal}
\end{figure*}

\vspace{-2mm}
\paragraph{Circadian activity.}
Figure~\ref{fig:temporal}(a) displays hourly tweet volume normalized within each community to account for corpus size differences. Because the corpus is not geolocated, we report hours in UTC and provide Gulf Standard Time (GST) conversions only as contextual approximations. BPD and Bipolar share a broadly similar circadian profile: activity troughs in the early morning UTC window (02:00 to 06:00 UTC; approximately 05:00 to 09:00 GST) and peaks in the afternoon/evening UTC window (BPD: 18:00 UTC; Bipolar: 21:00 UTC). ADHD differs from this pattern, exhibiting a midday UTC peak (11:00 UTC) with only 20.2\% of tweets falling in the 17:00 to 21:00 UTC window compared to 27.8\% for BPD and 33.1\% for Bipolar. All three communities share an early morning UTC trough (02:00 to 06:00 UTC).
BPD exhibits a gradual rise toward evening with moderate daytime activity; ADHD shows a midday-concentrated profile with greater hour-to-hour variability. Because no user-location validation is available, these results should be interpreted as platform-time activity patterns rather than direct evidence of Gulf-population temporal norms. Day-of-week peaks differ by community (Figure~\ref{fig:temporal}(b)): BPD on Monday, Bipolar on Sunday, ADHD on Friday.

 \vspace{-3mm} \paragraph{Longitudinal collection profile.}
Appendix Figure~\ref{fig:monthly} shows monthly tweet volume across the 2022 to 2026 collection window. The Bipolar community data span the full period; however, meaningful volume only emerges from 2024 onward (12 tweets before 2024 vs.\ 2,467 from 2024 to 2026), reflecting the expanding reach of the data collection pipeline rather than a gradual organic growth process. The BPD community is heavily concentrated in the 2025 collection window (February to October 2025), accounting for 83.5\% of its total volume in that period, with an additional 7.5\% from early 2026. ADHD data cover a narrower window (April 2024 to December 2025), with the majority of tweets from 2025 ($n$=211, 83.4\%), and are sparser overall, consistent with the smaller community sizes. The BPD temporal concentration introduces a confound: all BPD discourse patterns reported in this paper, vocabulary, cultural keyword rates, circadian patterns, and code-switching rates, are derived almost entirely from a single nine-month window and may reflect period-specific discourse rather than stable community characteristics. No date-stratified robustness check was performed; whether BPD findings replicate across different collection windows is unknown.
\vspace{-3mm}
\subsection{Linguistic and Cultural Framing}
\label{sec:linguistic}
\vspace{-2mm}
\paragraph{Community distinctive vocabulary via weighted log odds.}
To identify vocabulary statistically characteristic of each community, we apply the weighted log odds ratio with an informative Dirichlet prior \citep{monroe2008fightin}, a variance-normalized formulation that prevents rare words from dominating rankings through chance (full specification in Appendix~\ref{appendix:logodds}). Text is preprocessed by removing \texttt{[USER]} and \texttt{[URL]} placeholders, stripping non-Arabic characters, and filtering a standard Arabic stopword list. Words appearing fewer than five times in the target community are excluded.

Figure~\ref{fig:logodds} shows the top-10 distinctive terms per community ranked by standardized weighted log-odds ($z$) score, where higher positive $z$ values indicate words that are disproportionately associated with a given community relative to the remaining corpus. \textbf{The BPD community} is most strongly characterized by diagnostic terminology, including \ar{الحدية} (\textit{al-ḥadiyya}, ``borderline''; $z$=4.38) and \ar{الحدي} (\textit{al-ḥaddi}, ``borderline''; $z$=3.76), alongside identity vocabulary such as \ar{الشخصية} (\textit{al-shakhṣiyya}, ``personality''; $z$=4.10), relational terms including \ar{العلاقات} (\textit{al-ʿalāqāt}, ``relationships''; $z$=3.70), and affective vocabulary such as \ar{المشاعر} (\textit{al-mashāʿir}, ``feelings''; $z$=3.68) and \ar{الحب} (\textit{al-ḥubb}, ``love''; $z$=3.44). This pattern lexically overlaps with clinical constructs often discussed in relation to BPD, including interpersonal relationships, identity, and emotional regulation; however, lexical evidence alone cannot establish clinical features in users. As 83.5\% of BPD tweets originate from a single nine-month collection window (see Section~\ref{sec:temporal}), the temporal stability of this profile cannot be verified. \textbf{The Bipolar community} yields the highest absolute $z$ scores across all communities, led by \ar{القطب} (\textit{al-quṭb}, ``pole''; $z$=10.30) and \ar{ثنائي} (\textit{thunāʾī}, ``bi''; $z$=9.94), which together form the expression \textit{thunāʾī al-quṭb} (``bipolar''). Additional high-scoring terms include \ar{الله} (\textit{Allāh}, ``God/Allah''; $z$=9.79), \ar{الاكتئاب} (\textit{al-iktiʾāb}, ``depression''; $z$=7.83), \ar{الهوس} (\textit{al-hawas}, ``mania''; $z$=7.72), \ar{هوس} (\textit{hawas}, ``mania''; $z$=7.32), \ar{نوبه} (\textit{nawba}, ``episode''; $z$=6.14), and \ar{نوبة} (\textit{nawba}, ``episode''; $z$=6.00). Notably, \ar{الله} (\textit{Allāh}, ``God/Allah'') ranks third among the distinctive terms, aligning with the elevated religious keyword frequencies reported below. Collectively, the high $z$ scores suggest that discourse within the Bipolar community is characterized by a combination of condition-related, episodic, and religious vocabulary. \textbf{The ADHD community} shows a tighter $z$ score range (1.88 to 3.10), consistent with its smaller corpus reducing statistical power. The distinctive vocabulary is nonetheless semantically coherent: 
\ar{كونسيرتا} (\textit{Kūnsīrtā}, ``Concerta''; methylphenidate; $z$=3.10), \ar{الذكاء} (\textit{al-dhakāʾ}, ``intelligence''; $z$=2.92), \ar{الحركة} (\textit{al-ḥaraka}, ``movement/hyperactivity''; $z$=2.90), \ar{فرط} (\textit{farṭ}, ``excess/hyper-''; $z$=2.87), and \ar{الانتباه} (\textit{al-intibāh}, ``attention''; $z$=2.73).
\vspace{-3mm}
\paragraph{Latent topic structure via NMF.}
To characterize corpus-level lexical themes, we fit a non-negative matrix factorization (NMF; \citealp{lee1999learning}) topic model to TF--IDF representations of the self-disclosure-filtered tweets. The NMF pipeline removed very short reply fragments ($<$30 characters), stripped placeholders, normalized Arabic orthography, removed Arabic stopwords and community label terms, and used unigram TF--IDF features (min\_df=8, max\_df=0.70, max\_features=5,000). This left 7,192 tweets for the topic model, so the NMF analysis complements the full-corpus lexical and cultural-keyword analyses rather than replacing them. We selected the number of topics by searching $k=5$--14 using $C_v$ coherence; $k=12$ yielded the highest score ($C_v=0.5013$), although the improvement over $k=5$ ($C_v=0.4987$) was small. The resulting topics and top terms are listed in Appendix~\ref{appendix:lda} (Table~\ref{tab:nmf-topics}). The NMF results provide a view of community-level lexical structure. BPD shows the largest concentration in a feelings and relational-pain topic (Topic 1; 36.58\% of BPD tweets, versus 16.14\% of ADHD and 14.01\% of Bipolar), with additional BPD-weighted topics involving social support/advice (Topic 7; 6.63\%) and person/condition attribution (Topic 8; 7.77\%). Bipolar is comparatively elevated on an episode-vocabulary topic centered on mania and depression episodes (Topic 4; 14.48\%, versus 3.49\% for BPD and 3.59\% for ADHD), a treatment and medication topic (Topic 10; 22.33\%), and a gratitude/supplication topic (Topic 6; 7.80\%). ADHD, while still low-powered, is concentrated in first-person autobiographical framing (Topic 0; 19.73\%) and treatment/medication management (Topic 10; 20.18\%).\vspace{-3mm} \paragraph{Cultural keyword framework.}
We apply a dictionary-based approach using six Arabic keyword lists spanning religious, medical, family/social, emotional distress, identity, and stigma domains (Table~\ref{tab:cultural-keywords}, Appendix~\ref{appendix:cultural-keywords}), derived through iterative corpus review anchored to Kleinman's \citeyear{kleinman1980patients} explanatory models framework. The lists are an initial operationalization, not a validated instrument; rates reflect keyword prevalence, not direct measurements of latent constructs. All rates below are \textit{raw occurrence counts} per 100 tweets; the Religious Framing paragraph uses a more restricted list with binary tweet-level hit rates, and the two metrics are not directly comparable.
\vspace{-1mm}
Figure~\ref{fig:cultural-domains} shows keyword rates across all six domains; three findings are noteworthy. First, the Bipolar community displays the highest religious keyword rate (41.3 raw occurrences per 100 tweets), nearly 2.5 times that of BPD (16.7) and 2.1 times that of ADHD (19.8). This pattern is compatible with prior accounts of faith-based coping and meaning-making in Arabic mental health contexts \citep{eid2025somatic, zaghouani-etal-2026-posts}, though keyword counts alone cannot establish whether religiosity functions as coping, causation, routine pragmatic expression, or broader cultural discourse. Importantly, this finding should be read alongside the pipeline's documented under-sensitivity to indirect and metaphorical disclosure in the Bipolar community (the community with the lowest GPT-4.1 agreement against human gold, $\kappa \approx 0.49$; Section~\ref{sec:human-validation}). The reported rate of 41.3 may therefore underestimate religiously inflected disclosure, but the exact magnitude of any bias is unknown. Second, the Bipolar community also leads in medical keyword use (28.6 per 100 tweets), exceeding ADHD (24.1) and substantially exceeding BPD (11.7). To assess whether co-occurrence of religious and medical vocabulary reflects individual-level pluralism rather than two separate user populations, we computed the tweet-level intersection: \textbf{10.3\%} of Bipolar tweets (256 of 2,479) contain at least one keyword from both the religious and medical domains simultaneously, compared to 3.0\% for BPD (164 of 5,415) and 6.7\% for ADHD (17 of 253; this cell is too sparse for meaningful interpretation and is reported for completeness only). The Bipolar--BPD difference is statistically reliable ($\chi^2(1)=178.4$, $p<10^{-40}$), though we note this test addresses only the reliability of the contrast, not whether the keywords capture the constructs they are intended to measure. This tweet-level co-occurrence provides stronger, though still exploratory, evidence compatible with explanatory model pluralism \citep{kleinman1980patients} at the level of individual posts. Third, BPD exhibits the highest identity keyword rate (35.6 per 100 tweets) and emotional distress keyword rate (29.6), suggesting stronger lexical emphasis on selfhood and distress in this corpus. The ADHD community shows the lowest emotional distress keyword rate (12.3), suggesting a comparatively more practical or symptom-management-oriented discourse. All ADHD rates should be interpreted with lower confidence given the small subcorpus (see the ADHD statistical power note, Section~\ref{sec:analyses}). Family/Social and Stigma keyword rates are shown in Figure~\ref{fig:cultural-domains} for completeness; cross-community differences on these domains are smaller and are not among the three strongest signals in this corpus.
\vspace{-3mm}
\paragraph{Religious Framing.}
To examine religious language in greater depth, we applied a restricted religious keyword dictionary and organized matches into a four-tier exploratory taxonomy grounded in framing theory \citep{goffman1974frame} and interpreted through Kleinman's explanatory-model framework \citep{kleinman1980patients}. Unlike the broader religious inventory reported in Appendix~\ref{appendix:cultural-keywords}, which includes all religiously associated terms (including highly polysemous or pragmatically ambient expressions), the restricted analysis retains only terms that function as relatively unambiguous religious markers in mental health contexts, such as explicit supplication, Quranic references, and supernatural-causation vocabulary. 
This analysis uses a binary tweet-level metric in which each tweet contributes at most one match per tier. The taxonomy itself is researcher-defined and should therefore be interpreted as an exploratory analytical framework rather than a validated annotation scheme. Across the corpus, 15.3\% of tweets contain at least one keyword from the restricted analysis, with the highest binary hit rate observed in the Bipolar community (24.9 tweets per 100), followed by ADHD (13.8) and BPD (11.0).
The majority of religious tweets fall within \textit{(1) Ambient Expression} ($n=1{,}052$; 84.2\% of religious tweets), comprising culturally conventional expressions such as \ar{الحمد لله} (\textit{al-ḥamdu lillāh}, ``praise be to God'') and \ar{إن شاء الله} (\textit{in shāʾ Allāh}, ``God willing''), which frequently function as pragmatic discourse markers rather than illness-specific theological claims. A smaller category, \textit{(2) Coping \& Practice} ($n=104$; 9.1\%), includes references to prayer and Quranic recitation, such as \ar{الصلاة} (\textit{al-ṣalāh}, ``prayer'') and \ar{القرآن} (\textit{al-Qurʾān}, ``the Quran''), which may reflect faith-based coping practices or routine religious observance. More explicit moral and supernatural framing appears in \textit{(3) Guilt and Supernatural} ($n=61$; 4.9\%), which includes terms such as \ar{ذنب} (\textit{dhanb}, ``sin''), \ar{عقاب} (\textit{ʿiqāb}, ``punishment''), and \ar{الشيطان} (\textit{al-shayṭān}, ``Satan''). At the same time, 29 tweets (0.4\%) contain apparent counter-narratives, including expressions such as \ar{مو ذنبك} (\textit{mū dhanbak}, ``it is not your fault''), which explicitly reject blame-based attributions. The least frequent category, \textit{(4) Illness Causation Attribution} ($n=32$; 2.6\%), includes tweets that appear to frame mental health conditions in terms of divine trial, fate, or spiritual causation, including references such as \ar{ابتلاء} (\textit{ibtilāʾ}, ``divine trial''), \ar{القدر} (\textit{al-qadar}, ``fate/destiny''), and \ar{سببه روحي} (\textit{sababuhu rūḥī}, ``its cause is spiritual''). Taken together, the co-occurrence of medical and religious vocabulary within Bipolar discourse is compatible with explanatory-model pluralism; however, the present evidence remains dictionary-level and cannot establish users' underlying causal beliefs or treatment preferences. Overall, religious language in the corpus appears functionally heterogeneous: most instances consist of culturally ambient expressions, a smaller subset may reflect coping practices, and only a limited proportion encode explicit causal interpretations of mental health conditions (see Appendix~\ref{appendix:religious}). \vspace{-3mm} \paragraph{Code-switching.} To measure English code-switching, we removed \texttt{[USER]} and \texttt{[URL]} placeholder tokens from each tweet, then identified tweets containing at least one content-bearing English word, defined as a Latin-script token of two or more characters after excluding a standard English stopword list. Under this operationalization, \textbf{343} of \textbf{5,415} BPD tweets (\textbf{6.3}\%), \textbf{80} of \textbf{2,479} Bipolar tweets (\textbf{3.2}\%), and \textbf{72} of \textbf{253} ADHD tweets (\textbf{28.5}\%, 95\% CI: 22.9 to 34.1\%, binomial) contain content-bearing English.
The ADHD rate (28.5\%) is strikingly higher than BPD (6.3\%) and Bipolar (3.2\%). BPD English is dominated by diagnostic and therapeutic terminology: \textit{BPD} ($n$=79), \textit{DBT} (32), \textit{splitting} (20). Bipolar English is sparse, centering on condition labels (\textit{depression}, \textit{BD}). ADHD English is strongly anchored to the condition label itself (\textit{ADHD}, $n$=54) alongside neurodiversity-specific vocabulary (\textit{mindfulness}, \textit{Russell Barkley}). One plausible explanation is that the English acronym \textit{ADHD} functions as a compact, globally recognizable shorthand in online discourse, despite the availability of Arabic terminology for the condition \citep{alkhateeb2019adhd, alqahtani2025standardization}; this interpretation aligns with prior work on Arabic--English code-switching and bilingual lexical choice \citep{alamri2022arabic, myers1997duelling}, though user-level sociolinguistic factors cannot be ruled out.
\vspace{-2mm}
\section{Conclusion}
\label{sec:conclusion}
\vspace{-2mm}
We presented an exploratory computational characterization of Arabic mental health discourse across multiple condition-specific X Communities. Using a GPT-4.1-assisted personal-disclosure pipeline, we constructed a self-disclosure-filtered corpus of \textbf{8,147} tweets from \textbf{607} users and analyzed corpus composition, temporal activity, lexical distinctiveness, topic structure, code-switching, and cultural keyword prevalence. The results suggest community-associated patterns that extend beyond diagnostic vocabulary. In this corpus, Bipolar tweets show co-occurring religious and medical vocabulary, a pattern compatible with explanatory-model pluralism but insufficient to establish users' causal beliefs. Notably, \textbf{10.3}\% of Bipolar tweets contain keywords from both domains, providing a tweet-level signal more consistent with individual explanatory pluralism than aggregate community-level rates alone. BPD tweets foreground relational, identity, and emotional-distress vocabulary, whereas ADHD tweets more often center on practical symptom and medication management. These findings should therefore be interpreted as hypotheses for future work rather than confirmed condition-level properties. Future work should expand the ADHD corpus, validate the keyword and religious-framing schemes through inter-rater annotation, and extend this interpretable approach to additional conditions and Arabic dialect regions.

\section*{Limitations}
\label{sec:lim}
\textbf{Pipeline.} The prompt was developed on a Saudi-centric dataset, so adaptation may be needed for other Arabic dialects and regions. Inter-model agreement ($\kappa = 0.84$) overstates reliability relative to human-grounded estimates ($\kappa_{\text{GPT-human}} = 0.631$; $\kappa_{\text{Qwen-human}} = 0.329$). The conservative \textsc{Negative} default, bio override rule, and any-positive user aggregation may affect recall and precision, especially for indirect disclosures and highly active users.

\textbf{Corpus.} BPD data are temporally concentrated, with 83.5\% of tweets from a single nine-month window, and the ADHD subcorpus is small ($n$=253). The corpus is also not geolocated, and no dedicated bot-detection procedure was applied; temporal and community-level findings should therefore be interpreted cautiously.

\textbf{Analyses.} The cultural keyword framework and religious-framing taxonomy are exploratory rather than validated instruments. Keyword rates and the $\chi^2$ co-occurrence test support descriptive contrasts, but not construct validity. Future work should add user-level validation, dialect-sensitive keyword validation, robustness checks, and sensitivity analyses excluding GPT-positive/Qwen-negative users.

\section*{Ethical Considerations}
\label{sec:ethics}
 This study analyzes publicly visible posts from X Communities and does not involve
direct contact with users, recruitment, or intervention. Because the data concern
sensitive mental-health discourse, we treat the study as discourse-level analysis
rather than individual-level assessment. User identifiers were removed from the
analytic dataset prior to analysis, raw post text is not redistributed, and
results are reported only in aggregate.

Consistent with X content-redistribution restrictions, raw post text, usernames,
bios, profile metadata, user IDs, and community labels are not redistributed.
The public release provides only X Post IDs, which may be rehydrated by
authorized users through the X API subject to X's applicable Developer Agreement,
Developer Policy, access limits, and any required approvals. Because Post IDs can
be used to retrieve original posts when they remain available, the released data
should not be considered fully anonymized.

The \texttt{likely\_disclosure} label is an operational descriptor based on
self-reported experiential language and does not constitute a clinical diagnosis;
no clinical inferences should be drawn from pipeline outputs. The LLM-assisted
annotation pipeline may under-detect indirect, figurative, or culturally specific
forms of disclosure, particularly where religious or metaphorical language is
prevalent. The cultural keyword framework and religious-framing taxonomy are
exploratory instruments that have not been validated across Arabic dialect
regions. Findings should not be used to characterize, profile, screen, or
intervene on individual users, nor should aggregated discourse patterns be
generalized to Arabic-speaking populations or used to estimate the prevalence or
nature of mental-health conditions in Arab societies.
 
\bibliography{custom}

\appendix

\label{appendix}
 
\section{Personal Disclosure Classification Prompt}
\label{appendix:prompt}
 
This appendix presents the complete prompt used for LLM assisted personal disclosure classification of tweets from Arabic mental health communities. The prompt was run with GPT-4.1 as the primary annotator and Qwen3-235B-A22B as a screening model in parallel: GPT-4.1 (\texttt{GPT-4.1}, temperature=0.0, max\_tokens=250) and Qwen3-235B-A22B (\texttt{Qwen3-235B-A22B instruct 2507}, temperature=0.0, max\_tokens=250). Each model independently classifies each tweet as \textsc{Positive} (the tweet contains personal-disclosure signals: the author appears to be personally living with or experiencing a mental health condition) or \textsc{Negative} (no personal-disclosure signals), together with a confidence level and a set of reason tags. Classification operates at the tweet-level; the user bio is provided as supporting context. Table~\ref{tab:prompt-structure} summarizes the components.
 
\begin{table}[t]
\centering
\small
\begin{tabular}{@{}p{2.8cm}p{4.0cm}@{}}
\toprule
\textbf{Component} & \textbf{Function} \\
\midrule
System Role & Establishes domain expertise and task framing \\
Input Format & JSON object with community, bio, and tweet \\
Label Definitions & Positive/Negative criteria \\
Bio Override Rule & Bio driven Positive override and professional bio handling \\
Reason Tag Taxonomy & 13 tags covering disclosure signals, non-disclosure signals, bio signals, and edge cases \\
Confidence Levels & Three level confidence scale \\
Default Rule & Conservative Negative default \\
Few Shot Examples & 6 calibration exemplars (fully synthetic) \\
\bottomrule
\end{tabular}
\caption{Components of the personal disclosure classification prompt.}
\label{tab:prompt-structure}
\end{table}

\subsection{System Role and Task Definition}
\label{appendix:prompt-role}
 
\begin{promptbox}
\textbf{SYSTEM ROLE}
 
You are an expert classifier working with Arabic-language social media data collected from focused mental health communities on X (Twitter). These communities focus on ADHD, Bipolar Disorder, and BPD (Borderline Personality Disorder).
 
\medskip
\textbf{TASK}
 
Your task is to classify a single tweet as either:
\begin{itemize}[nosep,leftmargin=*]
  \item \texttt{positive}: the tweet contains personal-disclosure signals: the author appears to be personally living with or experiencing a mental health condition
  \item \texttt{negative}: the tweet contains no personal-disclosure signals: the content is educational, professional, neutral, or irrelevant
\end{itemize}
 
You are classifying the tweet only. User level decisions are handled separately.
\end{promptbox}

\subsection{Input Format and Bio Override Rule}
\label{appendix:prompt-input}
 
\begin{promptbox}
\textbf{INPUT FORMAT}
 
\begin{verbatim}
{
  "community": "ADHD | BPD | bipolar",
  "user_bio": "...",
  "tweet_text": "..."
}
\end{verbatim}
 
Use BOTH the bio and the tweet together to make your decision.
 
\medskip
\textbf{BIO OVERRIDE RULE}
 
If the bio clearly identifies the user as living with a condition, explicit diagnosis, living with language, or condition as personal identity (e.g., \ar{تم تشخيصي بـ $ADHD$}، \ar{أعيش مع ثنائي القطب}، \#BPD, ``bipolar girl''), then:
\begin{itemize}[nosep,leftmargin=*]
  \item Set \texttt{tweet\_label} to \texttt{positive}
  \item Add \texttt{BIO\_SELF\_IDENTIFICATION} to reason\_tags
  \item This applies even if the tweet itself contains no disclosure signal
\end{itemize}
 
If the bio indicates a professional or institutional account, this supports \texttt{negative}, but does NOT override a clearly personal tweet.
 
If the bio is empty, rely on the tweet alone and add \texttt{EDGE\_EMPTY\_BIO} to reason\_tags.
\end{promptbox}

\subsection{Classification Rules}
\label{appendix:prompt-labels}
 
\subsubsection{Classify as Positive}
\label{appendix:label-positive}
 
\begin{promptbox}
Classify as \textbf{positive} if ANY of the following are true:
 
\begin{itemize}[nosep,leftmargin=*]
  \item First person account of experiencing symptoms (e.g., \ar{أعاني من تشتت}، \ar{نفسيتي مدمرة}، \ar{ما أقدر أنام})
  \item Disclosing a personal diagnosis (e.g., \ar{تم تشخيصي بـ $ADHD$}، \ar{عندي ثنائي القطب})
  \item Sharing personal emotional distress or struggles related to a mental health condition
  \item Seeking peer support or venting about daily life with a condition
  \item Asking whether a personally experienced symptom belongs to a condition (first person question)
  \item Expressing experience from the inside, not explaining or educating others about a condition
\end{itemize}
\end{promptbox}
 
\subsubsection{Classify as Negative}
\label{appendix:label-negative}
 
\begin{promptbox}
Classify as \textbf{negative} if ANY of the following are true:
 
\begin{itemize}[nosep,leftmargin=*]
  \item Educational or psychoeducational content explaining symptoms or treatments in third person
  \item Written as a professional advising or answering someone else's question
  \item Promoting a therapy session, app, webinar, course, book, or product
  \item Discussing research findings, clinical definitions, or diagnostic criteria
  \item Addressing community members as a separate audience (e.g., \ar{هؤلاء الأشخاص يحتاجون...})
  \item Spam, off topic, or completely irrelevant content
\end{itemize}
 
\medskip
\textbf{Default Rule:} When in doubt, label \textbf{negative}.
\end{promptbox}

\subsection{Output Format, Reason Tag Taxonomy, and Confidence Levels}
\label{appendix:prompt-output}
 
\begin{promptbox}
\textbf{OUTPUT FORMAT}
 
Return ONLY a valid JSON object with no extra text, no markdown fences:
 
\begin{verbatim}
{
  "tweet_label": "positive | negative",
  "confidence": "high | medium | low",
  "reason_tags": ["TAG_1", "TAG_2"]
}
\end{verbatim}
\end{promptbox}
 
\begin{promptbox}
\textbf{REASON TAG TAXONOMY}
 
\medskip
\textit{Disclosure signal tags (support \texttt{positive}):}
\begin{itemize}[nosep,leftmargin=*]
  \item \texttt{TWEET\_SYMPTOM\_DISCLOSURE}: Tweet describes experiencing symptoms in first person
  \item \texttt{TWEET\_PERSONAL\_DIAGNOSIS\_DISCLOSURE}: Tweet explicitly states the user was diagnosed
  \item \texttt{TWEET\_PEER\_SUPPORT\_SEEKING}: Tweet seeks support or validation from others with the condition
  \item \texttt{TWEET\_EMOTIONAL\_VENTING}: Tweet expresses raw personal emotion or distress without educational intent
  \item \texttt{TWEET\_FIRST\_PERSON\_SYMPTOM\_QUESTION}: Tweet asks whether a personally experienced symptom belongs to a condition
\end{itemize}
 
\medskip
\textit{Non-disclosure signal tags (support \texttt{negative}):}
\begin{itemize}[nosep,leftmargin=*]
  \item \texttt{TWEET\_EDUCATIONAL\_CONTENT}: Tweet explains symptoms, conditions, or treatment in informational third person style
  \item \texttt{TWEET\_PROFESSIONAL\_ADVICE}: Tweet offers clinical guidance or answers someone else's question professionally
  \item \texttt{TWEET\_SERVICE\_PROMOTION}: Tweet promotes a therapy session, app, webinar, course, or mental health product
  \item \texttt{TWEET\_RESEARCH\_OR\_CLINICAL}: Tweet discusses diagnostic criteria, research findings, or clinical definitions
  \item \texttt{TWEET\_THIRD\_PERSON\_FRAMING}: Tweet addresses community members as a separate audience
  \item \texttt{TWEET\_SPAM\_OR\_IRRELEVANT}: Tweet is off topic, spam, or unrelated to mental health
\end{itemize}
 
\medskip
\textit{Bio signal tags (always added when detected, regardless of tweet label):}
\begin{itemize}[nosep,leftmargin=*]
  \item \texttt{BIO\_SELF\_IDENTIFICATION}: Bio clearly identifies the user as personally living with or diagnosed with a condition
\end{itemize}
 
\medskip
\textit{Edge case tags:}
\begin{itemize}[nosep,leftmargin=*]
  \item \texttt{EDGE\_EMPTY\_BIO}: Bio is absent; classification relies entirely on tweet content
  \item \texttt{EDGE\_AMBIGUOUS\_FIRST\_PERSON}: Tweet could be personal or professional; classified based on best available signal
  \item \texttt{EDGE\_PROFESSIONAL\_BIO\_}\allowbreak\texttt{PERSONAL\_TWEET}: Bio suggests a professional but tweet content is clearly personal/experiential
\end{itemize}
\end{promptbox}
 
\begin{promptbox}
\textbf{CONFIDENCE LEVELS}
 
\begin{itemize}[nosep,leftmargin=*]
  \item \texttt{high}: Strong unambiguous signal (e.g., explicit diagnosis disclosure, clear third person educational content)
  \item \texttt{medium}: Signal present but indirect or requires inference
  \item \texttt{low}: Very weak or contradictory signals; classification is a best guess
\end{itemize}
\end{promptbox}

\subsection{Important Notes}
\label{appendix:prompt-notes}
 
\begin{promptbox}
\begin{itemize}[nosep,leftmargin=*]
  \item This dataset is primarily in Arabic (Modern Standard and Gulf/Saudi dialect). Be sensitive to dialectal expressions of distress (e.g., \ar{مو زين}، \ar{نفسيتي مدمرة}، \ar{قرفانة من كل شي}).
  \item Do NOT base the classification solely on the community tag: professionals, researchers, and caregivers are present in all three communities.
  \item If the bio clearly identifies the user as living with a condition, always return \texttt{tweet\_label} \texttt{positive} and add \texttt{BIO\_SELF\_IDENTIFICATION}: even if the tweet itself is educational or neutral.
  \item A professional bio does NOT override a clearly personal tweet: classify such cases as \texttt{positive} and add \texttt{EDGE\_PROFESSIONAL\_BIO\_}\allowbreak\texttt{PERSONAL\_TWEET}.
  \item This classification is for research purposes. Handle all data with care and do not make clinical inferences beyond the binary label requested.
\end{itemize}
\end{promptbox}

\subsection{Few Shot Examples}
\label{appendix:prompt-examples}
 
All examples are \textbf{fully synthetic}; no real user data is reproduced. We present six exemplars covering representative label, confidence, and tag combinations. Table~\ref{tab:example-coverage} summarizes coverage.
 
\begin{table}[t]
\centering
\small
\begin{tabularx}{\columnwidth}{@{}lX@{}}
\toprule
\textbf{Dimension} & \textbf{Coverage (6 examples)} \\
\midrule
Label & Positive (3), Negative (3) \\
Confidence & High (4), Medium (2) \\
Bio signal tags & \texttt{BIO\_SELF\_IDENTIFICATION} (3) \\
Edge case tags & \small\texttt{EDGE\_EMPTY\_BIO} (1), \texttt{EDGE\_AMBIGUOUS\_}\allowbreak\texttt{FIRST\_PERSON} (1), \texttt{EDGE\_PROFESSIONAL\_BIO\_}\allowbreak\texttt{PERSONAL\_TWEET} (1) \\
\bottomrule
\end{tabularx}
\caption{Coverage of the six synthetic few-shot examples used in the prompt.}
\label{tab:example-coverage}
\end{table}
 
\paragraph{Example 1: Positive, high confidence (emotional venting + self-identifying bio).}
 
\begin{promptbox}
\noindent\textbf{Input:}\\
\texttt{community:} \texttt{"BPD"}\\
\texttt{user\_bio:} \ar{إنسانة تتعلم كيف تعيش مع اضطراب الشخصية الحدية يومًا بيوم} \texttt{| \#BPD}\\
\texttt{tweet\_text:} \ar{أصعب شي في الحدية إنك تحب بشكل كامل وفجأة تحس إن كل شي انهار بدون سبب واضح}\\[2pt]
 
\textbf{Output:} \texttt{\{"tweet\_label": "positive", "confidence": "high", "reason\_tags": ["TWEET\_EMOTIONAL\_VENTING", "BIO\_SELF\_IDENTIFICATION"]\}}
\end{promptbox}
 
\paragraph{Example 2: Negative, high confidence (educational content + third person framing).}
 
\begin{promptbox}
\noindent\textbf{Input:}\\
\texttt{community:} \texttt{"ADHD"}\\
\texttt{user\_bio:} \ar{أخصائي نفسي إكلينيكي , ماجستير إرشاد نفسي , مرخص من هيئة التخصصات الصحية}\\
\texttt{tweet\_text:} \ar{الفرق بين فرط الحركة عند الأطفال والبالغين: الأطفال يُظهرون أعراضًا حركية واضحة، بينما يعاني البالغون من أعراض داخلية كالقلق الذهني وصعوبة التنظيم.}\\[2pt]
 
\textbf{Output:} \texttt{\{"tweet\_label": "negative", "confidence": "high", "reason\_tags": ["TWEET\_EDUCATIONAL\_CONTENT", "TWEET\_THIRD\_PERSON\_FRAMING"]\}}
\end{promptbox}
 
\paragraph{Example 3: Positive, high confidence (symptom disclosure + first person question, empty bio).}
 
\begin{promptbox}
\noindent\textbf{Input:}\\
\texttt{community:} \texttt{"bipolar"}\\
\texttt{user\_bio:} \texttt{""}\\
\texttt{tweet\_text:} \ar{أحس هالأيام بطاقة زايدة عن اللزوم، ما أنام، وعندي رغبة أشتري أشياء ما أحتاجها ! هل هذا طبيعي ولا ممكن يكون هوس؟}\\[2pt]
 
\textbf{Output:} \texttt{\{"tweet\_label": "positive", "confidence": "high", "reason\_tags": ["TWEET\_SYMPTOM\_DISCLOSURE", "TWEET\_FIRST\_PERSON\_SYMPTOM\_QUESTION", "EDGE\_EMPTY\_BIO"]\}}
\end{promptbox}
 
\paragraph{Example 4: Negative, medium confidence (professional bio, ambiguous tweet).}
 
\begin{promptbox}
\noindent\textbf{Input:}\\
\texttt{community:} \texttt{"BPD"}\\
\texttt{user\_bio:} \ar{معالج نفسي معتمد , متخصص في اضطرابات الشخصية , باحث في العلاج الجدلي السلوكي $DBT$}\\
\texttt{tweet\_text:} \ar{أحيانًا الشفاء لا يبدو كالشفاء ! بل يبدو كلحظة هدوء صغيرة وسط العاصفة. أتمنى لكم تلك اللحظة}\\[2pt]
 
\textbf{Output:} \texttt{\{"tweet\_label": "negative", "confidence": "medium", "reason\_tags": ["TWEET\_THIRD\_PERSON\_FRAMING", "EDGE\_AMBIGUOUS\_FIRST\_PERSON"]\}}
\end{promptbox}
 
\paragraph{Example 5: Positive, high confidence (educational tweet, bio self-identification override).}
 
\begin{promptbox}
\noindent\textbf{Input:}\\
\texttt{community:} \texttt{"ADHD"}\\
\texttt{user\_bio:} \ar{مبرمج ومهتم بالتقنية , تم تشخيصي بـ $ADHD$ منذ سنتين}\\
\texttt{tweet\_text:} \ar{الفرق بين فرط الحركة عند الأطفال والبالغين من وجهة نظر علمية}\\[2pt]
 
\textbf{Output:} \texttt{\{"tweet\_label": "positive", "confidence": "high", "reason\_tags": ["TWEET\_EDUCATIONAL\_CONTENT", "BIO\_SELF\_IDENTIFICATION"]\}}
\end{promptbox}
 
\paragraph{Example 6: Positive, medium confidence (emotional venting + symptom disclosure, professional bio override).}
 
\begin{promptbox}
\noindent\textbf{Input:}\\
\texttt{community:} \texttt{"bipolar"}\\
\texttt{user\_bio:} \ar{طالبة دكتوراه علم نفس , أعيش مع ثنائي القطب وأحاول أفهمه من الداخل والخارج}\\
\texttt{tweet\_text:} \ar{لما تكون في نوبة اكتئاب وتعرف نظريًا كل الأدوات العلاجية بس ما تقدر تطبق ولو واحدة ! هذا تناقض ما يفهمه غير اللي عاشه}\\[2pt]
 
\textbf{Output:} \texttt{\{"tweet\_label": "positive", "confidence": "medium", "reason\_tags": ["TWEET\_EMOTIONAL\_VENTING", "TWEET\_SYMPTOM\_DISCLOSURE", "BIO\_SELF\_IDENTIFICATION", "EDGE\_PROFESSIONAL\_BIO\_PERSONAL\_TWEET"]\}}
\end{promptbox}

\subsection{User Level Aggregation}
\label{appendix:prompt-aggregation}
 
Tweet level classifications are aggregated into a single user-level label using a deterministic priority ordered procedure. 
Each user is assigned one of two labels: \textsc{likely\_disclosure} or \textsc{other}. The rules are applied in order; the first matching rule determines the outcome.
 
\begin{enumerate}[nosep]
  \item \textbf{Bio override (AGG\_BIO\_DISCLOSURE\_OVERRIDE).}
    If \emph{any} tweet for the user carries \texttt{BIO\_SELF\_IDENTIFICATION} in its \texttt{reason\_tags}, the user is immediately labeled \textsc{likely\_disclosure}, regardless of tweet-level labels. This rule fires because the bio override in the tweet-level prompt propagates the same tag to every tweet for that user; detecting it once is sufficient.
 
  \item \textbf{Any positive tweet wins (AGG\_ANY\_POSITIVE\_TWEET / AGG\_CONFLICT\_POSITIVE\_WINS).}
    If any tweet-level label is \texttt{positive}, the user is labeled \textsc{likely\_disclosure}.
    The aggregation reason distinguishes two sub cases:
    \begin{itemize}[nosep,leftmargin=*]
      \item \texttt{AGG\_ANY\_POSITIVE\_TWEET}: all tweets are positive (unanimous)
      \item \texttt{AGG\_CONFLICT\_POSITIVE\_WINS}: at least one tweet is positive but at least one is negative (conflict resolved in favor of positive)
    \end{itemize}
    The triggering tweet IDs are recorded for traceability.
 
  \item \textbf{All negative, no bio signal (AGG\_ALL\_NEGATIVE\_NO\_BIO\_SIGNAL).}
    If no tweet is positive and no bio signal was detected, the user is labeled \textsc{other}.
\end{enumerate}
 
The design of Rules~1 and~2 prioritizes recall for personal-disclosure evidence; this choice may increase false positives and should be considered when interpreting the corpus. Rule~3 provides the default for users whose discourse is entirely non-personal-disclosure. 
 
\section{Prompt Design Rationale}
\label{appendix:rationale}
 
\subsection{Theoretical Framework}
 
The prompt's core design decisions are grounded in three complementary frameworks. \textbf{\citet{goffman1959presentation} presentation of self.} The bio is the user's ``front stage'' presentation; tweets represent ``back stage'' behavior. This justifies reading both together and allowing a self-identifying bio to override a neutral tweet, while a clearly personal tweet overrides a professional bio signal.
 
 \textbf{The explanatory models} in   \citet{kleinman1980patients}. The distinction between personal experience of illness and third party or professional discourse about illness aligns with Kleinman's separation of illness experience from disease frameworks, motivating the Positive/Negative boundary and the tweet-level classification rules.
 
\textbf{Code-switching theory \citep{myers1997duelling}.} The important notes section's explicit sensitivity to Gulf Arabic and Saudi dialect expressions of distress ensures that dialectally expressed disclosures are recognized regardless of linguistic form.
 
The two label scheme reflects a single analytically motivated boundary: whether the tweet contains personal-disclosure signals suggesting the author is personally living with or experiencing a mental health condition, versus any other stance (professional, educational, or irrelevant). The structured reason tag taxonomy, covering disclosure signal tags, non-disclosure signal tags, bio signal tags, and edge case tags, provides a multi dimensional audit trail that makes the basis of each classification transparent and supports systematic error analysis.
 
\subsection{Conservative Default and Bio Override Rules}
 
Two design choices are especially consequential. First, the \textbf{Negative default} ensures that ambiguous tweets, content that discusses mental health generally, provides educational information, or addresses community members in the third person, are not mistakenly counted as personal-disclosure discourse. This is analytically conservative: some genuine disclosure tweets may be lost, but the resulting corpus is less contaminated by non-disclosure content. Second, the \textbf{bio override} rule, a self-identifying bio triggers a Positive label regardless of tweet content, captures users who may post educational or neutral content on a given day while living with a condition. Conversely, a professional bio does not override a clearly personal tweet, preventing the systematic exclusion of clinicians or researchers who also share their own lived experience.

\section{Dataset Statistics}
\label{appendix:dataset}
 
\begin{table}[t]
\centering
\small
\begin{tabular}{@{}lr@{}}
\toprule
\textbf{Statistic} & \textbf{Value} \\
\midrule
Total tweets (post preprocessing) & 9,582 \\
Total tweets (post removal) & \textbf{8,147} \\
Unique users (post preprocessing) & 1,286 \\
Unique users (post removal) & \textbf{607} \\
Communities & 3 communities\\
 & across 3 conditions \\
 
& (see Table~\ref{tab:communities}) \\
\midrule
\multicolumn{2}{@{}l}{\textit{Bio status ($\leq$\,2 words = MINIMAL)}} \\
\quad FULL ($\geq$\,3 words) & 790 (61.4\%) \\
\quad MINIMAL ($\leq$\,2 words) & 152 (11.8\%) \\
\quad EMPTY (null/blank) & 344 (26.7\%) \\
\midrule
\multicolumn{2}{@{}l}{\textit{User label distribution (pipeline output)}} \\
\quad LIKELY\_DISCLOSURE & \textbf{607} (47.2\%) \\
\quad OTHER & \textbf{679} (52.8\%) \\
\bottomrule
\end{tabular}
\caption{Dataset statistics. EMPTY = null/blank bio; MINIMAL = $\leq$\,2 words; FULL = $\geq$\,3 words.}
\label{tab:dataset-stats}
\end{table}

\clearpage
\section{Annotation Quality and Validation}
\label{appendix:validation}
 
\subsection{Inter Model Agreement Protocol}
 
GPT-4.1 and Qwen3-235B-A22B were run on the same prompt across all 9,582 preprocessed tweets. Both models received identical inputs (community, user\_bio, tweet\_text) and produced independent \textsc{Positive}/\textsc{Negative} labels. Cohen's $\kappa$ was computed across all tweet pairs where neither model returned a parse failure. GPT-4.1 returned 20 partial parses (0.2\%) and zero full failures; Qwen3 returned 111 parse failures (1.2\%). Raw agreement on valid pairs ($n = 9{,}471$): 90.8\%; Cohen's $\kappa = 0.84$.
 
\textbf{Interpretation.} This inter-model $\kappa$ should not be read as a standalone validity claim. As shown in the human validation study below, it overstates pipeline reliability because GPT-4.1 and Qwen3 share a conservative bias on easy cases while diverging sharply on ambiguous ones. Qwen3's primary function is to flag disagreements for review, not to serve as an independent validator.
 
\subsection{Human Validation Study}
 
\subsubsection{Sample and annotators}
Two native Arabic-speaking annotators independently labeled a stratified sample of 200 tweets, all of which had valid paired labels. The sample was stratified across five difficulty tiers: high confidence positive (\textit{C\_pos\_high}, $n$=50), low/medium confidence positive (\textit{B\_pos\_low\_med}, $n$=30), high confidence negative (\textit{E\_neg\_high}, $n$=30), low/medium confidence negative (\textit{D\_neg\_low\_med}, $n$=30), and inter-model conflict (\textit{A\_conflict}, $n$=60). Each tweet was labeled \textsc{Positive} or \textsc{Negative} using the same guidelines as the LLM prompt.
 
\subsubsection{Inter human agreement}
The annotators agreed on 192 of 200 tweets (96.00\%), with $\kappa = 0.905$ (almost perfect). All 8 disagreements were directionally consistent: one annotator applied a more liberal threshold on ambiguous positive cases. The 192 agreed tweets constitute the human gold standard used to evaluate the LLMs.
 
\subsubsection{LLM performance against human gold}
Table~\ref{tab:iaa} summarizes GPT-4.1 and Qwen3 performance against the 192 agreed human gold labels. GPT-4.1 reaches substantial agreement ($\kappa = 0.631$; F$_1^+ = 0.88$); Qwen3 reaches only fair agreement ($\kappa = 0.329$; F$_1^+ = 0.72$), confirming the inter-model $\kappa = 0.84$ overstates pipeline reliability.
 
\begin{table}[t]
\centering
\small
\resizebox{\columnwidth}{!}{%
\begin{tabular}{@{}lcccc@{}}
\toprule
\textbf{Comparison} & \textbf{$\kappa$} & \textbf{Agree.} & \textbf{F$_1^+$} & \textbf{Band} \\
\midrule
Inter human (ceiling) & 0.905 & 96.00\% & NA & Almost perfect \\
GPT-4.1 vs.\ human gold & 0.631 & 83.85\% & 0.88 & Substantial \\
Qwen3 vs.\ human gold & 0.329 & 66.30\% & 0.72 & Fair \\
\midrule
GPT-4.1 vs.\ Qwen3 & 0.84 & 90.80\% & NA & Almost perfect \\
\bottomrule
\end{tabular}%
}
\caption{Human validation results ($n$=200). F$_1^+$: positive-class F$_1$. GPT-4.1 evaluated on 192 gold labels; Qwen3 on 184 (after parse failures). Inter-model row shown for reference only.}
\label{tab:iaa}
\end{table}
 
\subsubsection{Stratum level breakdown}
Table~\ref{tab:stratum_agreement} shows per-stratum agreement. GPT-4.1 performs well on clear-label strata (C\_pos\_high: 96\%; E\_neg\_high: 93\%; B\_pos\_low\_med: 90\%) but drops to 46\% on low/medium confidence negatives (D\_neg\_low\_med), where the conservative \textsc{Negative} default suppresses genuine disclosures. Qwen3 collapses on inter-model conflict cases (18\%), confirming it cannot serve as an independent annotator on ambiguous tweets.
 
\begin{table}[t]
\centering
\small
\begin{tabular}{@{}lcc@{}}
\toprule
\textbf{Stratum} & \textbf{GPT agree.} & \textbf{Qwen3 agree.} \\
\midrule
C\_pos\_high ($n$=50) & 96\% & 96\% \\
B\_pos\_low\_med ($n$=30) & 90\% & 90\% \\
E\_neg\_high ($n$=28) & 93\% & 93\% \\
D\_neg\_low\_med ($n$=26) & 46\% & 46\% \\
A\_conflict & 83\% ($n$=58) & 18\% ($n$=50) \\
\bottomrule
\end{tabular}
\caption{Per-stratum agreement with human gold. $n$ values reflect the subset of each stratum with mutual human agreement (192 total), after excluding the 8 human-disagreement tweets from the full 200-tweet sample. D\_neg\_low\_med is the critical failure stratum; Qwen3 collapses on conflict cases (18\%).}
\label{tab:stratum_agreement}
\end{table}
 
\subsubsection{Community breakdown (GPT-4.1 vs.\ human gold)}
Agreement is highest for ADHD ($\kappa = 0.73$), followed by BPD ($\kappa = 0.66$), and lowest for Bipolar ($\kappa \approx 0.49$). Bipolar accounts for 10 of 21 GPT false negatives on the human gold set, consistent with that community's indirect, religious, and metaphorical disclosure style being most susceptible to the conservative \textsc{Negative} default.

\section{Supplementary Figures and Formulas}
\label{appendix:lda}
 
\subsection{Log-Odds Formulation}
\label{appendix:logodds}
 
For a target group $i$ and comparison group $j$, the log-odds of word $w$ under the weighted log-odds ratio with an informative Dirichlet prior \citep{monroe2008fightin} is:
\begin{multline}
\delta^{(i-j)}_w = \log \frac{y^i_w + \alpha_w}{n^i + \alpha_0 - y^i_w - \alpha_w} \\
- \log \frac{y^j_w + \alpha_w}{n^j + \alpha_0 - y^j_w - \alpha_w}
\end{multline}
where $y^i_w$ is the count of word $w$ in group $i$, $n^i$ is the total word count of group $i$, and $\alpha_w$ is the prior count drawn from the pooled background corpus ($\alpha_0 = \sum_w \alpha_w$). The $z$ score normalizes by variance:
\begin{equation}
\zeta^{(i-j)}_w = \frac{\delta^{(i-j)}_w}{\sqrt{\sigma^2_w}}, \quad \sigma^2_w = \frac{1}{y^i_w + \alpha_w} + \frac{1}{y^j_w + \alpha_w}
\end{equation}
 
The background prior $\alpha_w$ is set to the word's frequency in the full corpus; unseen words receive additive smoothing of 0.01.
 
\subsection{NMF Topic Model}
 
\begin{table*}[t]
\centering
\scriptsize
\resizebox{\textwidth}{!}{%
\begin{tabular}{@{}clp{7.8cm}rccc@{}}
\toprule
\textbf{ID} & \textbf{Human label} & \textbf{Top terms} & \textbf{$n$} & \textbf{BPD \%} & \textbf{Bipolar \%} & \textbf{ADHD \%} \\
\midrule
0  & First-person narrative & \ar{اني، كنت، اقدر، اعرف، وانا، اقول، حياتي، لاني، ابدا، اكون، افكر، وما} & 522 & 6.86 & 6.86 & 19.73 \\
1  & Feelings and relational pain & \ar{بل، المشاعر، الشخص، شعور، الحدي، احيانا، ليس، الشعور، دون، الالم، الحب، الخوف} & 2105 & 36.58 & 14.01 & 16.14 \\
2  & Questions and peer validation & \ar{هل، ام، سوال، طبيعي، علاقه، الحديين، عليكم، ليه، بالحديه، وهل، السوال، تحسون} & 361 & 4.44 & 6.30 & 5.38 \\
3  & Clinical encounters and causes & \ar{لي، بالنسبه، سبب، وانا، صار، الدكتور، يسبب، يوم، قالت، شهور، فيني، تقريبا} & 593 & 8.03 & 8.84 & 7.17 \\
4  & Bipolar episode vocabulary & \ar{نوبه، الهوس، الاكتياب، هوس، اكتياب، وقت، نوبات، النوبه، مختلطه، طبيعي، تجيني، قلبي} & 485 & 3.49 & 14.48 & 3.59 \\
5  & Self-struggle/help seeking & \ar{نفسي، يوم، احاول، اكره، لاني، اكون، الناس، اخصايي، تعبت، ليش، احتاج، اسوي} & 604 & 8.78 & 7.33 & 10.31 \\
6  & Gratitude and supplication & \ar{شكرا، يارب، اللهم، امين، خير، فعلا، عليك، كلامك، شاء، يسعدك، ربي، العافيه} & 306 & 2.79 & 7.80 & 2.24 \\
7  & Advice and social support & \ar{نفسك، عليك، الا، محد، راح، اهم، معك، نصيحه، وانت، حاول، حياتك، منك} & 407 & 6.63 & 3.62 & 4.04 \\
8  & Person/condition attribution & \ar{شخص، حدي، طبيعي، الشخص، راح، عنده، يعاني، معي، لانه، مصاب، واحد، نفسه} & 471 & 7.77 & 3.95 & 4.93 \\
9  & Uncertainty and diffuse distress & \ar{احس، مدري، ناس، حولي، تعبت، الشي، دايما، ذا، ليش، ليه، مشاعري، الناس} & 306 & 4.58 & 3.57 & 3.59 \\
10 & Treatment and medication & \ar{الادويه، العلاج، السلوكي، علاج، بعض، الجدلي، النفسي، ادويه، الحمدلله، اخذ، النفسيه، الدكتور} & 927 & 8.41 & 22.33 & 20.18 \\
\bottomrule
\end{tabular}%
}
\caption{NMF topic labels, top terms, topic sizes, and within-community percentages. The model was selected by $C_v$ coherence over $k=5$--14; $k=12$ produced the highest coherence ($C_v=0.5013$). Labels are researcher-assigned and exploratory.}
\label{tab:nmf-topics}
\end{table*}
  
\section{Limitations and Dialect Extensibility}
\label{appendix:extensibility}
The classification prompt was developed and validated on a Saudi-centric dataset. The important notes section explicitly flags sensitivity to Gulf Arabic and Saudi dialect expressions of distress, but the few shot examples and the implicit discourse norms encoded in the classification rules skew Gulf Arabic. Table~\ref{tab:extensibility} identifies areas requiring adaptation for other Arabic dialect groups.
 
\begin{table}[t]
\centering
\small
\begin{tabular}{@{}p{1.6cm}p{5.0cm}@{}}
\toprule
\textbf{Dialect} & \textbf{Required Adaptations} \\
\midrule
Egyptian & Indirect or figurative disclosure patterns (e.g., \ar{زهقت من نفسي}); colloquial symptom vocabulary \\
Levantine & Reflexive expressions for personal suffering (\ar{حالي}/\ar{حالتي}); different medication brand names \\
N.\ African & French Arabic code-switching in clinical and symptom vocabulary \\
\midrule
\multicolumn{2}{@{}l}{\textit{Cross dialect}} \\
\midrule
Indirect disclosure & Users who describe experience metaphorically or religiously without explicit diagnosis language may be under detected by the explicit disclosure criterion \\
Condition labels & ADHD has multiple Arabic renderings and frequent English-acronym use in online discourse; other conditions may have competing translations \\
\bottomrule
\end{tabular}
\caption{Dialect specific and cross dialect adaptations needed beyond Gulf Arabic / MSA.}
\label{tab:extensibility}
\end{table}

\section{Religious Framing Analysis}
\label{appendix:religious}
 
\begin{figure*}[t]
    \centering
    \includegraphics[width=0.8\textwidth]{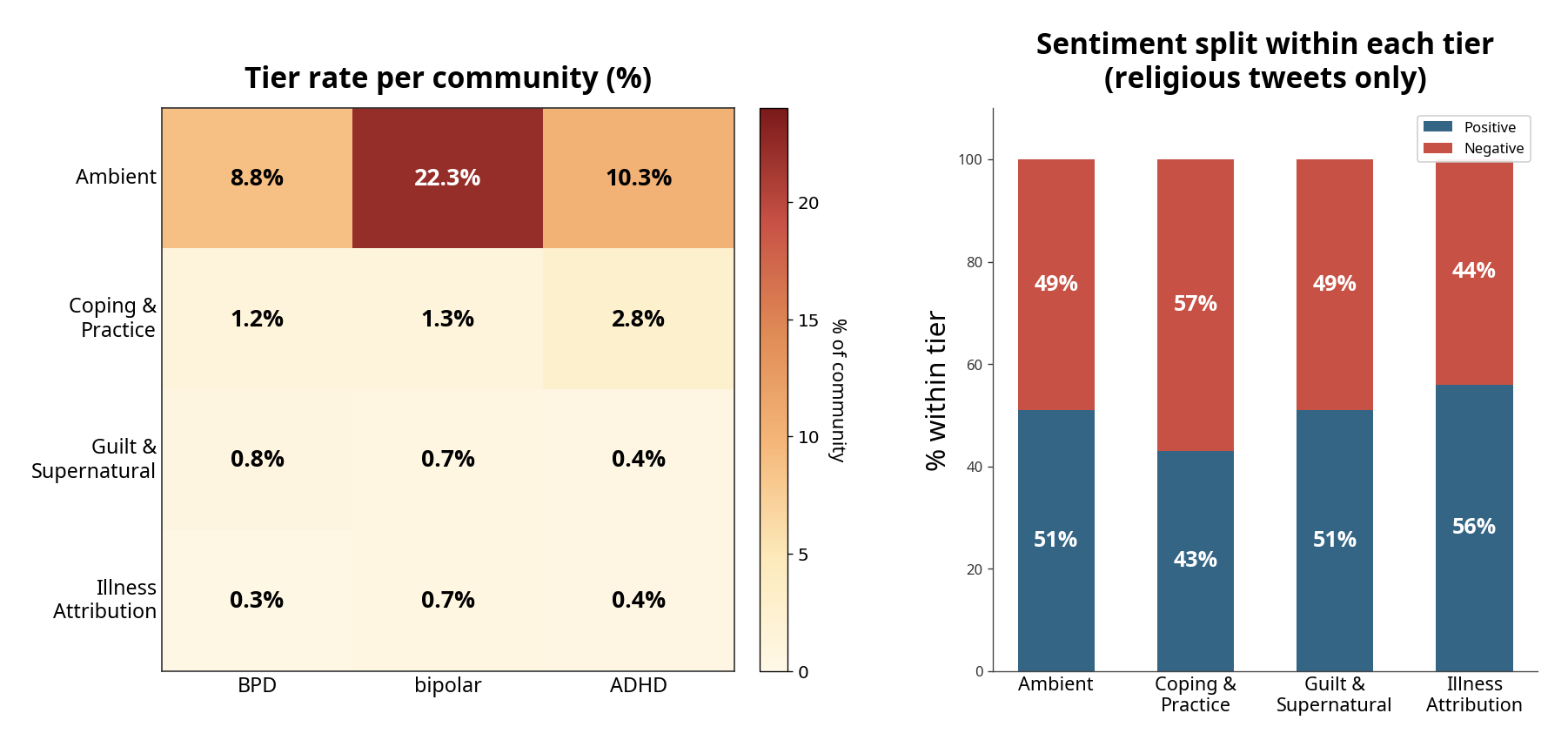}
    \caption{Religious framing tier rates per community (left) and sentiment split within each tier (right). Cell values = \% of community tweets containing a tier keyword; sentiment computed over religious-language tweets only.}
    \label{fig:r1}
\end{figure*}
 
The religious analysis defines a four-tier taxonomy of language use: \textit{Ambient Expression}, \textit{Coping \& Practice}, \textit{Guilt \& Supernatural}, and \textit{Illness Causation Attribution}.
\paragraph{Tier Distribution and Within-Tier Sentiment}
Figure~\ref{fig:r1} shows that \textit{Ambient Expression} dominates across communities, accounting for 8.8\% (BPD) to 22.3\% (Bipolar) of total tweet volume, with the Bipolar rate more than 2.5$\times$ BPD and roughly double ADHD (10.3\%), reinforcing earlier evidence of elevated religious framing; this estimate is likely conservative due to under-detection in Bipolar discourse. The remaining tiers each account for under 3\% of tweets but are analytically important: \textit{Coping \& Practice} peaks in ADHD (2.8\%), reflecting symptom-practice interplay, while \textit{Guilt \& Supernatural} and \textit{Illness Causation Attribution} are most concentrated in Bipolar (0.7\% each), consistent with explanatory-model pluralism. Sentiment patterns differ by tier: \textit{Ambient Expression} is near-balanced (51\% positive, 49\% negative), \textit{Coping \& Practice} skews negative (57\%), \textit{Illness Causation Attribution} shows the strongest positive skew (56\%), and \textit{Guilt \& Supernatural} remains near-balanced, indicating that morally and supernaturally framed language does not straightforwardly align with negative affect.

\begin{figure*}[t]
    \centering
    \includegraphics[width=0.9\textwidth]{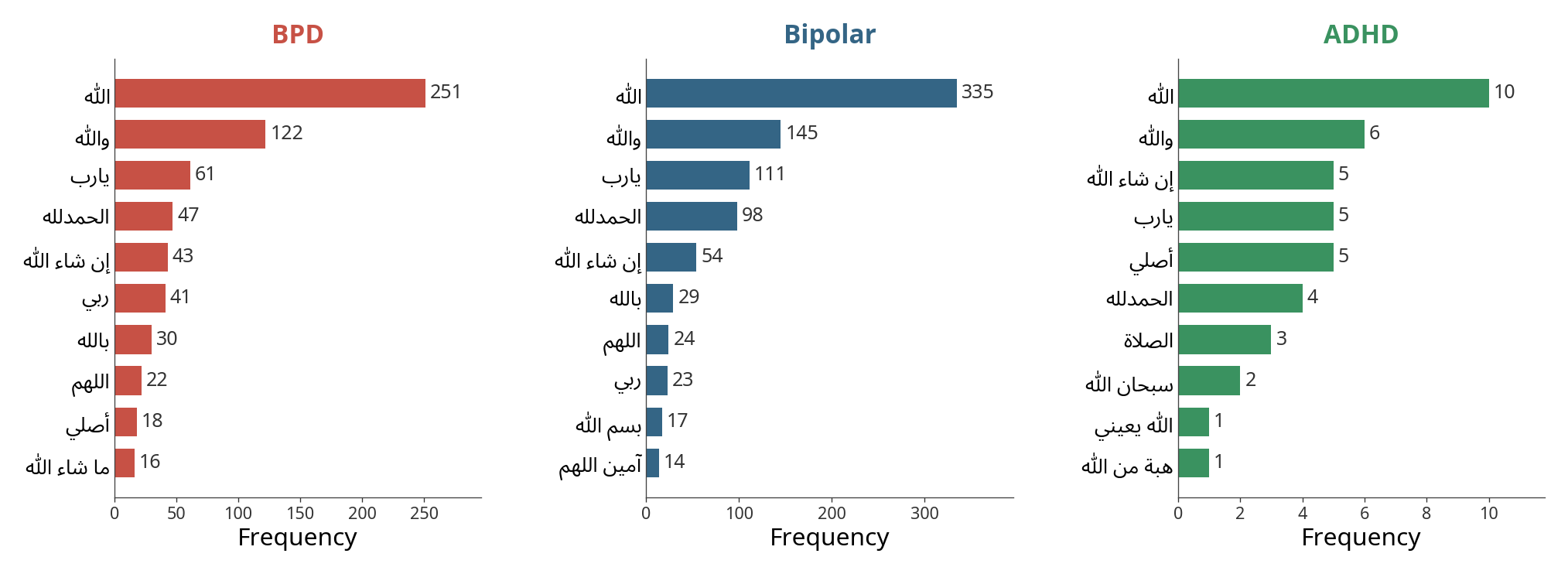}
    \caption{Top-10 religious keywords per community (ranked by raw count). Note the scale difference: BPD and Bipolar counts reach hundreds; ADHD reaches a maximum of 10 ($n$=253 tweets).}
    \label{fig:r2}
\end{figure*}

\begin{figure*}[t]
  \centering
  \includegraphics[width=\textwidth]{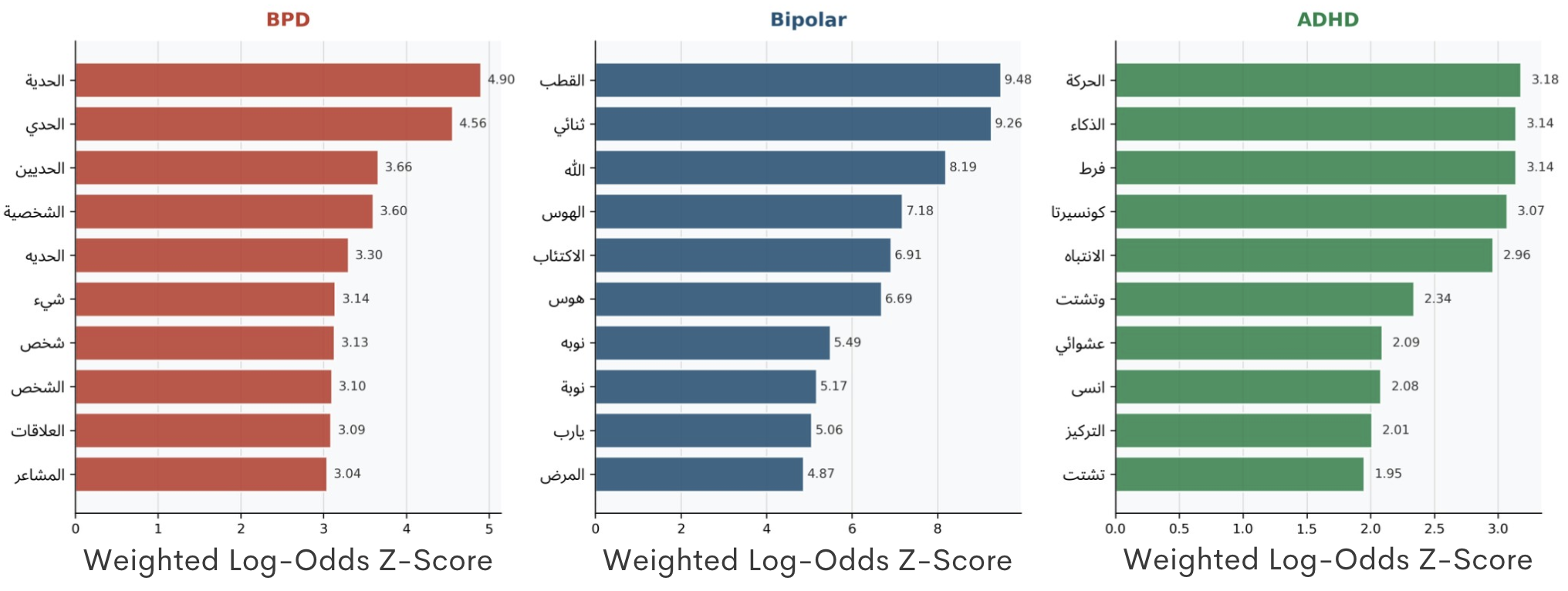}
  \caption{Top-10 community-distinctive words by weighted log-odds $z$-score. BPD: relational/diagnostic; Bipolar: episode/religious; ADHD: symptom/medication.
  }
  \label{fig:logodds}
\end{figure*}

\begin{figure*}[t]
  \centering
  \includegraphics[width=1\textwidth]{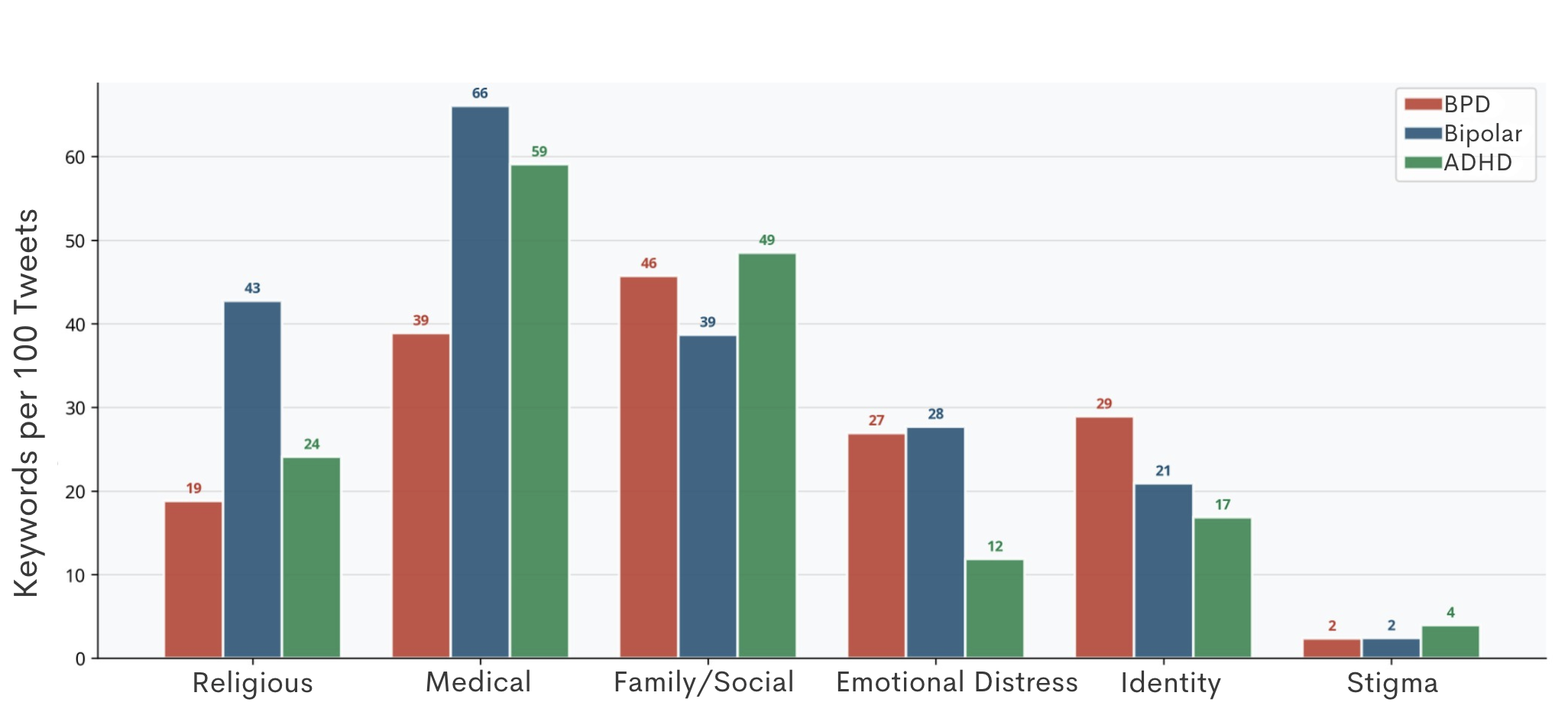}
  \caption{Cultural domain keyword rates (per 100 tweets). Bipolar leads on Religious and Medical; BPD on Identity and Emotional Distress.
  }
  \label{fig:cultural-domains}
\end{figure*}

\begin{figure*}[t]
  \centering
  \includegraphics[width=1\textwidth]{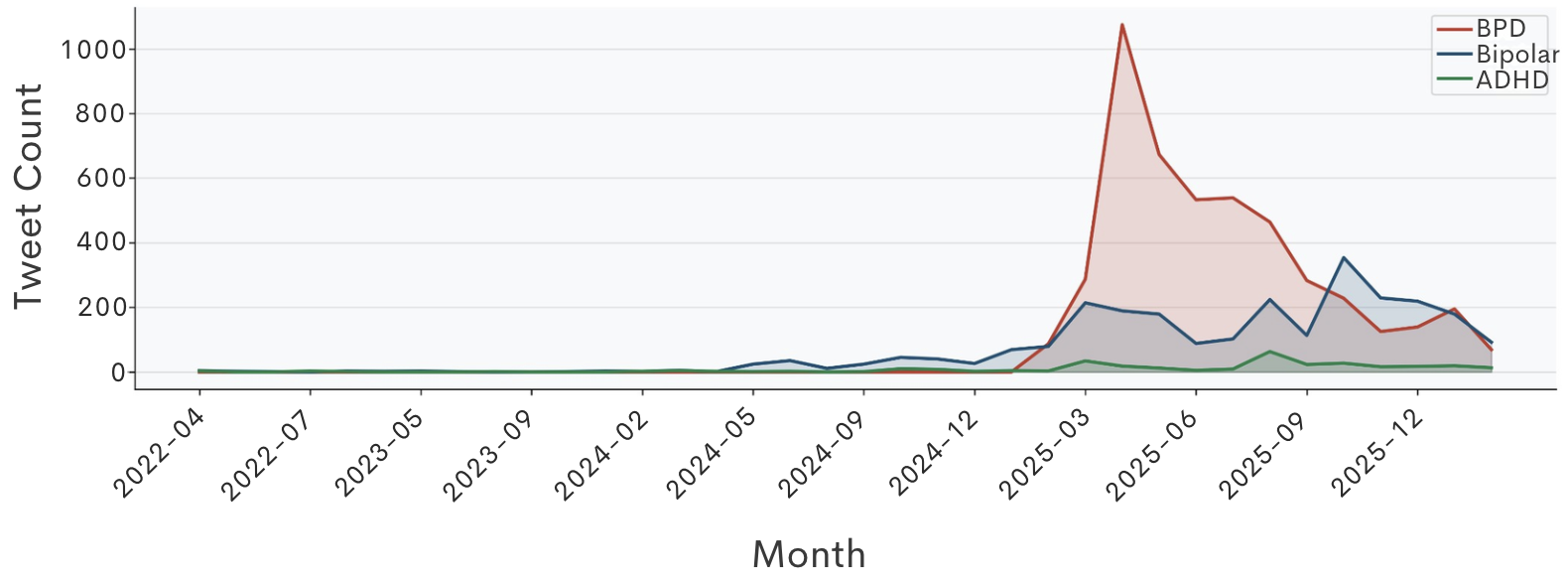}
  \caption{Monthly tweet volume (2022--2026). BPD concentrated in 2025; Bipolar spans the full period but with meaningful volume from 2024 onward.
  }
  \label{fig:monthly}
\end{figure*}
\vspace{-3mm}
\paragraph{Top Religious Keywords per Community}
Figure \ref{fig:r2} compares the most frequent religious keywords across BPD, Bipolar, and ADHD communities. Across all three, \ar{الله} (“God/Allah”) is the most frequent term, followed by \ar{والله} (“by God”) and \ar{يارب} (“O Lord”), indicating a shared reliance on core religious expressions in mental health discourse.
The Bipolar and BPD communities show similar patterns, with frequent use of \ar{الحمدلله} and \ar{إن شاء الله}. The Bipolar community additionally exhibits higher use of more formal supplicatory expressions such as \ar{بسم الله} and \ar{آمين اللهم}. The BPD community is distinct in its use of \ar{ما شاء الله}, often occurring in peer-support contexts.
In contrast, the ADHD community shows lower overall frequencies but includes more references to prayer-related terms such as \ar{أصلي} and \ar{الصلاة}, suggesting greater emphasis on ritual practice and routine. Overall, while all communities share common religious expressions, their usage differs in function across coping, peer interaction, and practice-oriented framing.
\section{Cultural Keyword Framework}
\label{appendix:cultural-keywords}
Table~\ref{tab:cultural-keywords} lists the six cultural keyword domains and representative Arabic terms used in the dictionary-based analysis (Section~\ref{sec:linguistic}). Each domain was derived through iterative corpus vocabulary review anchored to Kleinman's \citeyear{kleinman1980patients} explanatory models framework. The lists are an initial operationalization and have not yet been validated through an independent inter-rater annotation study.
 
\begin{table}[h]
\centering
\footnotesize
\begin{tabular}{@{}lp{4.6cm}@{}}
\toprule
\textbf{Domain} & \textbf{Example Keywords (Arabic)} \\
\midrule
Religious     & \ar{الله، ربي، يارب، صبر، الشفاء، شاء} \\
Medical       & \ar{دواء، دكتور، تشخيص، مستشفى، جلسة} \\
Family/Social & \ar{أهل، ماما، بابا، زوج، أصدقاء، مجتمع} \\
Emot.\ Distress & \ar{حزن، خوف، قلق، تعب، وحيد، اكتئاب} \\
Identity      & \ar{أنا، نفسي، شخصيتي، ذاتي، إحساسي} \\
Stigma        & \ar{مجنون، خبل، عيب، خجل، انكار} \\
\bottomrule
\end{tabular}
\caption{Cultural keyword domains and representative Arabic terms.}
\label{tab:cultural-keywords}
\end{table}
 
\section{X Community Overview}
\label{appendix:communities}
 
Table~\ref{tab:communities} summarizes the three X Communities used for corpus construction, including total membership size and moderation type at the time of data collection.
 
\begin{table}[h]
\centering
\small
\begin{tabular}{lrl}
\hline
\textbf{Condition} & \textbf{\# Members} & \textbf{Moderation} \\
\hline
BPD & 5,876 & Professional \\
ADHD & 2,321 & Varies \\
Bipolar & 1,320 & Peer led \\
\hline
\textbf{Total} & \textbf{9,517} & \\
\hline
\end{tabular}
\caption{X Communities used for corpus construction. \textit{Professional}: licensed psychologists; \textit{Varies}: mixed moderation; \textit{Peer-led}: lived-experience individuals or relatives.}
\label{tab:communities}
\end{table}

\end{document}